# Solution of FPK Equation for Stochastic Dynamics Subjected to Additive Gaussian Noise via Deep Learning Approach

Amir H. Khodabakhsh[1], Seid H. Pourtakdoust[2*]

[1] *Department of Aerospace Engineering, Sharif University of Technology, Tehran, Iran.*
*khodabakhsh@ae.sharif.edu*, ORCID: 0000-0002-0457-8673

[2] *Department of Aerospace Engineering, Sharif University of Technology, Tehran, Iran.*
*pourtak@sharif.edu*, ORCID: 0000-0001-5717-6240

## Abstract

The Fokker-Plank-Kolmogorov (FPK) equation is an idealized model representing many stochastic systems commonly encountered in the analysis of stochastic structures as well as many other applications. Its solution thus provides an invaluable insight into the performance of many engineering systems. Despite its great importance, the solution of the FPK equation is still extremely challenging. For systems of practical significance, the FPK equation is usually high dimensional, rendering most of the numerical methods ineffective. In this respect, the present work introduces the FPK-DP Net as a physics-informed network that encodes the physical insights, i.e. the governing constrained differential equations emanated out of physical laws, into a deep neural network. FPK-DP Net is a mesh-free learning method that can solve the density evolution of stochastic dynamics subjected to additive white Gaussian noise without any prior simulation data and can be used as an efficient surrogate model afterward. FPK-DP Net uses the dimension-reduced FPK equation. Therefore, it can be used to address high-dimensional practical problems as well. To demonstrate the potential applicability of the proposed framework, and to study its accuracy and efficacy, numerical implementations on five different benchmark problems are investigated.

---

[*] Corresponding Author: Tel: (+98) 21 6602 2731, Email: pourtak@sharif.edu

# Solution of FPK Equation for Stochastic Dynamics Subjected to Additive Gaussian Noise via Deep Learning Approach

## Introduction

Stochastic dynamical systems are often encountered in practical engineering and scientific problems [1]. The inherent uncertainties associated with stochastic dynamics in general and stochastic structures, in particular, render deterministic approaches for trajectory derivation impossible [2]. However, one can study the characteristics of these random responses by analyzing the evolution of the Probability Density Function (PDF). The Fokker-Plank-Kolmogorov (FPK) equation is one of the commonly utilized tools to model the evolution of a dynamical system PDF with stochastic input [3]. As an idealized model for numerous phenomena in various scientific disciplines, the FPK equation quantifies and captures the probability density evolution of stochastic systems toward a wide range of applications [3, 4]. An estimation of the probability density distribution for a stochastic structure can be leveraged to improve the robustness of its behavior or even increase safety while maintaining the desired level of reliability [5].

Due to its practical importance, many studies have been dedicated to solving the FPK equation over the past several decades [3, 6]. The exact analytical solution to the FPK equation can only be obtained under rigorous simplifying assumptions [4]. In general, the exact solution can be obtained if either (1) the drift is linear and the diffusion is constant, (2) both the drift and diffusion obey certain potential conditions, or (3) the separation of the variables is possible [4]. Numerical methods can be used. However, the spatial domain for the FPK equation is unbounded which is sometimes challenging to be implemented in numerical schemes.

Approximate solutions, even for a well-posed spatial boundary condition, are not straightforward either [7]. In addition, for practical applications, the FPK equation is usually high dimensional. The curse of dimensionality, in this case, makes most numerical grid-based methods intractable [6]. Monte Carlo (MC) method is another numerical approach used to find an approximate solution. However, MC is computationally intensive even for toy problems [8]. Owing to these difficulties, a generic solution to the FPK equation is still an open problem [3, 9].

With the recent surge in parallel computing power and significant advances in deep learning methods, the scientific community is now utilizing Machine Learning (ML) techniques to tackle complex classical problems. This shift of paradigm has shown great promise in a variety of applications, such as natural language processing [10] and image recognition [11]. Physics-informed learning is one such idea that was first proposed to solve differential equations back in the 1990s [12, 13]. However, the learning tools were not as powerful then to show the great potential of these networks. Recent progress in Automatic Differentiation (AD) techniques [14] and the development of powerful ML software packages such as TensorFlow [15], PyTorch [16], and Theano [17] have once again revived the idea of physics-informed learning by the scientific community.

In a recent study, Raissi et al. used this idea and proposed Physics-Informed Neural Networks (PINNs) [18]. For any physically grounded problem, generally the physical laws deduce to a set of nonlinear time-dependent Partial Differential Equations (PDEs) with any perceivable limitations. Forcing the neural network to satisfy these mathematical models (PDEs) ensures that the predictions comply with any conservation principle or invariances required. Utilizing AD, DNNs can learn to satisfy any differential operators defined over the inputs and outputs while representing the input-output relation [18, 19]. Unlike most numerical techniques, this idea does not require any mesh generation that in most cases, is a cumbersome and time-consuming task.

Conventional DNNs usually require a large set of labeled data to be trained. These labeled data may not be available or be expensive to collect. On the other hand, physics-informed DNNs do not require any prior labeled training data. The loss function in conventional supervised learning is defined based on the error between the computed output of the network and some training/test data. However, in physics-informed networks, the physical insight on some dynamics (the known governing differential equations, either partial or ordinary) is used to encode the physical restriction in the network for training purposes. To this end, the differential equation, in its residual form, is used as the loss function. In a classical sense, the solution satisfies the differential equation pointwise. Hence, the domain is sampled randomly, and the norm of the pointwise residuals is considered as the loss function to train the network. Ideally, if the network represents the exact solution, the norm of the pointwise residuals vanishes, and the loss function becomes zero. Otherwise, one can optimize the network's weights and biases to minimize the loss function. If the training is successful, the network represents the solution of the differential equation.

Xu et al. proposed the idea of solving the steady-state FPK equation using a physics-informed DNN [20]. They utilized a normalization condition to avoid a trivial solution. They also investigated the effects of the network depth, the penalty factors, and the optimization algorithm. In another study, Uy and Grigoriu used the same idea via a reformulation of the FPK equation to an integrodifferential equation. They considered the transient response [7]. Although these research studies showed a successful implementation of the physics-informed DNN to solve the FPK equation, they also noted that as the dimension of the problem increases, so does the complexity of the training process. In this paper, the high dimension complexity issue is addressed. In particular, a novel physics-informed DNN framework is proposed to solve the FPK equation. Solving the probability density evolution provides invaluable insight into the behavior of stochastic structures and their performance. However, for most applications of practical significance, the FPK equation that governs the time evolution of the probability density is high-dimensional, and numerical determination of the density evolution is a formidable task. To this end, a decoupling scheme is suggested that helps to handle the training process more efficiently. Utilizing a decoupling scheme lets one break the problem into several smaller sub-problems that are easier to manage. One of the main advantages of the DNNs is their parallel architecture that is utilized the most in this study. The current authors have recently developed DeepPDEM [21], as a novel concept based on the physics-informed deep neural networks [18] that solve the Generalized Density Evolution Equation (GDEE) [22]. Utilizing the Probability Density Evolution Method (PDEM) and the decoupled equivalent form of the FPK equation, a physics-informed DNN called the FPK-DP Net is proposed to solve the density evolution problem. In the following sections, the decoupled equivalent form of the FPK equation is explained. Subsequently, the equivalent formulation is used to define the FPK-DP Net.

## Decoupled Equivalent of the FKP Equation

Suppose that over some probability space $(\Omega, \mathcal{F}, P)$, an Itô process is describing the state of a general stochastic dynamic system $\{\mathbf{X}_t\}_{t \in T} \in \mathbb{R}^n$, via the following equation:

$$d\mathbf{X}_t = \mathbf{f}(t, \mathbf{X}_t)dt + \mathbf{g}(t, \mathbf{X}_t)d\mathrm{B}_t \quad t \geq 0 \tag{1}$$

Or equivalently,

$$\dot{\mathbf{X}}_t = \mathbf{f}(t, \mathbf{X}_t) + \mathbf{g}(t, \mathbf{X}_t)\mathbf{W}_t(\omega) \quad t \geq 0 \tag{2}$$

where $f: \mathbb{R}^{n+1} \to \mathbb{R}^n$ represents the drift vector, $g: \mathbb{R}^{n+1} \to \mathbb{R}^{n \times m}$ is known as the diffusion matrix, $\mathbf{B}_t$ is a zero mean ($E[d\mathrm{B}_t] = 0$) m-dimensional Brownian process vector, and $\mathbf{W}_t(\omega) = \frac{d\mathbf{B}_t}{dt}$ is an m-dimensional white noise process [4]. Let $\wp_\mathbf{X}(\mathbf{x}) \geq 0$ be the joint probability distribution of the process $\mathbf{X}_t$. Assuming additive Brownian process excitation, the evolution of the probability density $\wp_\mathbf{X}(\mathbf{x})$ for $\mathbf{X}_t$ is governed by the Fokker-Plank-Kolmogorov (FPK) equation [23].

$$\frac{\partial \wp_\mathbf{X}(\mathbf{x},t)}{\partial t} = -\nabla_\mathbf{x} \cdot \big(\mathbf{f}(\mathbf{x},t)\wp_\mathbf{X}(\mathbf{x},t)\big) + \mathbf{D}(t) : \mathcal{H}_\mathbf{x}\big(\wp_\mathbf{X}(\mathbf{x},t)\big) \tag{3}$$

where ":" is the Frobenius inner product (double-contraction product) and $\mathcal{H}_\mathbf{x}(\cdot)$ is the Hessian matrix. The diffusion $\mathbf{D}$ is a positive-definite symmetric tensor defined as:

$$\mathbf{D} = \frac{1}{2}\mathbf{g}\mathbf{g}^T \quad ; \quad D_{ij} = \frac{1}{2}\sum_{k=1}^{m} g_{ik} g_{jk} \tag{4}$$

The initial conditions might be deterministic which can be modeled as Dirac's delta distribution. On the other hand, there might be uncertainty in the initial condition that can be modeled as an initial probability distribution, $\wp_{\mathbf{X}_0}(\mathbf{x}_0)$. The initial distribution is either derived by applying statistical analysis over a set of empirical data (frequentist interpretation) or is inferred via some physical laws and prior experiments (Bayesian interpretation) [2, 24]. Without loss of generality, let the initial density distribution, $\wp_{\mathbf{X}_0}(\mathbf{x}_0)$, be known and independent of the stochastic input.

$$\wp_\mathbf{X}(\mathbf{x}, t_0) = \wp_{\mathbf{X}_0}(\mathbf{x}_0) \tag{5}$$

For natural boundary conditions, it is reasonable to assume that the probability density vanishes at infinity sufficiently fast, ensuring the normalization integral $\int_{-\infty}^{\infty} \wp_\mathbf{X}(\mathbf{x},t) d\mathbf{x} = 1$ (2nd axiom of the probability).

$$\lim_{|x_i| \to \infty} \wp_\mathbf{X}(\mathbf{x}, t) = 0 \tag{6}$$

As stated in [3], defining the probability flux (current) $\mathring{\jmath}$, we can rewrite Eq. (3) as a conservation equation, where the probability current or flux is defined as:

$$\mathring{\jmath}(\mathbf{x}, t) = (\mathbf{f}(\mathbf{x}, t) - \mathbf{D}\nabla_\mathbf{x})\wp_\mathbf{X}(\mathbf{x}, t) \tag{7}$$

Therefore, assuming that $\partial \wp_\mathbf{X}(\mathbf{x}, t)/\partial x_i$ exists for all $x_i$, the probability flux for the $i^{th}$ state is:

$$\mathring{\jmath}_i(\mathbf{x}, t) = f_i(\mathbf{x}, t)\wp_\mathbf{X}(\mathbf{x}, t) - \sum_{j=1}^{n} D_{ij} \frac{\partial \wp_\mathbf{X}(\mathbf{x}, t)}{\partial x_j} \tag{8}$$

From the continuity equation, presuming that $\partial \mathring{\jmath}_i(\mathbf{x}, t)/\partial x_i$ exists for all $x_i$, we have:

$$\frac{\partial \wp_\mathbf{X}(\mathbf{x}, t)}{\partial t} + \nabla_\mathbf{x}\mathring{\jmath}(\mathbf{x}, t) = 0 \tag{9}$$

Given the joint PDF $\wp_\mathbf{X}(\mathbf{x}, t)$, one can find the marginal PDF of $X_i$ as,

$$\wp_{X_i}(x_i, t) = \int_{\mathbb{R}^{n-1}} \wp_\mathbf{X}(\mathbf{x}, t) d\mathbf{v} \quad \text{where} \quad \mathbf{v} = \mathbf{x} \setminus \{x_i\} \tag{10}$$

Given the natural boundary condition, both sides of Eq. (10) can be differentiated with respect to time, or in other words:

$$\frac{\partial \wp_{X_i}(x_i, t)}{\partial t} = \int_{\mathbb{R}^{n-1}} \frac{\partial}{\partial t} \wp_\mathbf{X}(\mathbf{x}, t) d\mathbf{v} \tag{11}$$

Substituting Eq.(9) into Eq.(11) yields:

$$\frac{\partial \wp_{X_i}(x_i, t)}{\partial t} = -\int_{\mathbb{R}^{n-1}} \nabla_\mathbf{x}\mathring{\jmath}(\mathbf{x}, t) d\mathbf{v} \tag{12}$$

Since $\lim_{x_j \to \pm\infty} \mathring{\jmath}_j(\mathbf{x}, t) = 0$, we have $\int_{-\infty}^{\infty} \frac{\partial \mathring{\jmath}_j(\mathbf{x}, t)}{\partial x_j} dx_j = 0$. Thus,

$$\int_{\mathbb{R}^{n-1}} \nabla_\mathbf{x}\mathring{\jmath}(\mathbf{x}, t) d\mathbf{v} = \int_{\mathbb{R}^{n-1}} \frac{\partial \mathring{\jmath}_i(\mathbf{x}, t)}{\partial x_i} d\mathbf{v} \tag{13}$$

Substituting Eq. (13) into Eq. (12) gives:

$$\frac{\partial p_{X_i}(x_i, t)}{\partial t} = -\int_{\mathbb{R}^{n-1}} \frac{\partial \mathbb{j}_i(\mathbf{x}, t)}{\partial x_i} d\mathbf{v} \qquad (14)$$

Now assuming:

I. $\mathbb{j}_i(\mathbf{x}, t)$ is a Lebesgue-integrable function of $\mathbf{v}$ for each $x_i \in \mathbb{R}$,
II. for almost all $\mathbf{v} \in \mathbb{R}^{n-1}$, the derivative $\frac{\partial \mathbb{j}_i(\mathbf{x},t)}{\partial x_i}$ exists for all $x_i \in \mathbb{R}$,
III. for almost every $\mathbf{v} \in \mathbb{R}^{n-1}$ and for all $x_i \in \mathbb{R}$ there exists an integrable function $\phi: \mathbb{R}^{n-1} \to \mathbb{R}$ such that $\left|\frac{\partial \mathbb{j}_i(\mathbf{x},t)}{\partial x_i}\right| \leq \phi(\mathbf{v})$,

from the Leibniz integral rule, one can write:

$$\frac{\partial p_{X_i}(x_i, t)}{\partial t} = -\frac{\partial}{\partial x_i} \int_{\mathbb{R}^{n-1}} \mathbb{j}_i(\mathbf{x}, t) d\mathbf{v} \qquad (15)$$

Let $\mathbb{J}_i = \int_{\mathbb{R}^{n-1}} \mathbb{j}_i(\mathbf{x}, t) d\mathbf{v}$ denote the total flux of the probability in the $i^{th}$-direction. Thus, Eq. (15) can be written as:

$$\frac{\partial p_{X_i}(x_i, t)}{\partial t} = -\frac{\partial \mathbb{J}_i}{\partial x_i} \qquad (16)$$

Thus, for the total probability flux in the $i^{th}$-direction, we will have:

$$\mathbb{J}_i = \int_{\mathbb{R}^{n-1}} \mathbb{j}_i(\mathbf{x}, t) d\mathbf{v} = \int_{\mathbb{R}^{n-1}} f_i(\mathbf{x}, t) p_{\mathbf{X}}(\mathbf{x}, t) d\mathbf{v} - \int_{\mathbb{R}^{n-1}} \sum_{j=1}^{n} D_{ij} \frac{\partial p_{\mathbf{X}}(\mathbf{x}, t)}{\partial x_j} d\mathbf{v} \qquad (17)$$

Let $\mu_i = \int_{\mathbb{R}^{n-1}} f_i(\mathbf{x}, t) p_{\mathbf{X}}(\mathbf{x}, t) d\mathbf{v}$ and $\sigma_i = \int_{\mathbb{R}^{n-1}} \sum_{j=1}^{n} D_{ij} \frac{\partial p_{\mathbf{X}}(\mathbf{x},t)}{\partial x_j} d\mathbf{v}$, therefore, $\mathbb{J}_i = \mu_i - \sigma_i$.

Again, since the PDF vanishes at infinity ($\int_{-\infty}^{\infty} \frac{\partial p_{\mathbf{X}}(\mathbf{x},t)}{\partial x_k} dx_k = p_{\mathbf{X}}(\infty, t) - p_{\mathbf{X}}(-\infty, t) = 0$), $\sigma_i$ can be written as:

$$\sigma_i = \int_{\mathbb{R}^{n-1}} \sum_{j=1}^{n} D_{ij} \frac{\partial p_{\mathbf{X}}(\mathbf{x}, t)}{\partial x_j} d\mathbf{v} = D_{ii} \int_{\mathbb{R}^{n-1}} \frac{\partial p_{\mathbf{X}}(\mathbf{x}, t)}{\partial x_i} d\mathbf{v} \qquad (18)$$

Imposing the same assumptions on $p_{\mathbf{X}}(\mathbf{x}, t)$ as $\mathbb{j}_i(\mathbf{x}, t)$, using the Leibniz integral rule, and also noting that $\int_{\mathbb{R}^{n-1}} p_{\mathbf{X}}(\mathbf{x}, t) d\mathbf{v} = p_{X_i}(\mathbf{x}, t)$, $\sigma_i$ can be further simplified into:

$$\sigma_i = D_{ii} \int_{\mathbb{R}^{n-1}} \frac{\partial \wp_{\mathbf{X}}(\mathbf{x},t)}{\partial x_i} d\mathbf{v} = D_{ii} \frac{\partial}{\partial x_i} \int_{\mathbb{R}^{n-1}} \wp_{\mathbf{X}}(\mathbf{x},t) d\mathbf{v} = D_{ii} \frac{\partial \wp_{X_i}(x_i,t)}{\partial x_i} \qquad (19)$$

where $\sigma_i$ represents the diffusion effect and is now decoupled in terms of $X_i$. In what follows, the Probability Density Evolution Method (PDEM) [25] is utilized to decouple the drift term $\mu_i$ in terms of $X_i$. Once again, consider the governing Stochastic Differential Equation (SDE) given in Eq. (2). Utilizing a decomposition scheme, such as the Kosambi–Karhunen–Loève (KKL) theorem or Stochastic Harmonic Function (SHF) representation [26], the Wiener process in Eq.(2) can be decomposed into an infinite linear combination of orthogonal functions [25].

$$\dot{\mathbf{X}}_t = \mathbf{f}(t, \mathbf{X}_t) + \mathbf{g}(t, \mathbf{X}_t) \mathbf{h}(\mathbf{\Theta}, t) \quad t \geq 0 \qquad (20)$$

where $h(\cdot)$ is a real-valued continuous function and $\mathbf{\Theta}$ is an uncorrelated random variable vector with support $\Omega_{\mathbf{\Theta}}$ and joint PDF $\wp_{\mathbf{\Theta}}(\mathbf{\theta})$. To find the evolution of the density for each state $X_i$, the Generalized Density Evolution Equation (GDEE) [27] can be utilized.

$$\frac{\partial \tilde{\wp}_{X_i \mathbf{\Theta}}(x_i, \mathbf{\theta}, t)}{\partial t} + \dot{X}_i(\mathbf{\theta}, t) \frac{\partial \tilde{\wp}_{X_i \mathbf{\Theta}}(x_i, \mathbf{\theta}, t)}{\partial x_i} = 0 \qquad (21)$$

Solving (21) gives $\tilde{\wp}_{X_i \mathbf{\Theta}}(x_i, \mathbf{\theta}, t)$, which can be integrated over $\Omega_{\mathbf{\Theta}}$ to find $\tilde{\wp}_{X_i}(x_i, t)$.

$$\tilde{\wp}_{X_i}(x_i, t) = \int_{\Omega_{\mathbf{\Theta}}} \tilde{\wp}_{X_i \mathbf{\Theta}}(x_i, \mathbf{\theta}, t) d\mathbf{\theta} \qquad (22)$$

Since both the GDEE and FPK equations represent the same density evolution, the equivalence yields [23]:

$$\int_{\mathbb{R}^{n-1}} f_i(\mathbf{x}, t) \wp_{\mathbf{X}}(\mathbf{x}, t) d\mathbf{v} = \int_{\Omega_{\mathbf{\Theta}}} f_i(\mathbf{X}(\mathbf{\theta}, t), t) \tilde{\wp}_{X_i \mathbf{\Theta}}(x_i, \mathbf{\theta}, t) d\mathbf{\theta} \qquad (23)$$

From Eq.(20) we have:

$$f_i\big(t, X_i(\boldsymbol{\theta}, t)\big) = \dot{X}_i(\boldsymbol{\theta}, t) - \sum_{j=1}^{m} \Big(g_{ij}\big(t, X_i(\boldsymbol{\theta}, t)\big) h_j(\boldsymbol{\theta}, t)\Big) \tag{24}$$

Assuming that $\tilde{p}_{X_i\boldsymbol{\Theta}}(x_i, \boldsymbol{\theta}, t)$ and $\dot{X}_i(\boldsymbol{\theta}, t)$ are available by solving the GDEE, substituting Eq. (24) into Eq. (23) yields:

$$\mu_i = \int_{\Omega_{\boldsymbol{\Theta}}} \left( \dot{X}_i(\boldsymbol{\theta}, t) - \sum_{j=1}^{m} \Big(g_{ij}\big(t, X_i(\boldsymbol{\theta}, t)\big) h_j(\boldsymbol{\theta}, t)\Big) \right) \tilde{p}_{X_i\boldsymbol{\Theta}}(x_i, \boldsymbol{\theta}, t) d\boldsymbol{\theta} \tag{25}$$

Eq. (25) leads to the equivalent one-dimensional FPK equation. It is evident that $\tilde{p}_{X_i}(x_i, t)$ can be obtained directly from the GDEE. However, previous studies show that solving the FPK equation provides better results [9]. On the other hand, the marginal PDF found by the GDEE can be leveraged to reformulate the dimension-reduced equation into an FPK-like PDE for dynamic systems involving randomness in both parameters and excitation [9, 28]. To this end, recall the FPK equation (Eq. (9)) and let $\tilde{f}_i(x_i, t)$ that is referred to as the equivalent drift coefficient, to be:

$$\mu_i = \tilde{f}_i(x_i, t) \tilde{p}_{X_i}(x_i, t) \tag{26}$$

Or equivalently,

$$\tilde{f}_i(x_i, t) = \frac{\mu_i}{\tilde{p}_{X_i}(x_i, t)} \tag{27}$$

that can be approximated using the GDEE equation as:

$$\tilde{f}_i(x_i, t) = \frac{\int_{\Omega_{\boldsymbol{\Theta}}} \left( \dot{X}_i(\boldsymbol{\theta}, t) - \sum_{j=1}^{m} \Big(g_{ij}\big(t, X_i(\boldsymbol{\theta}, t)\big) h_j(\boldsymbol{\theta}, t)\Big) \right) \tilde{p}_{X_i\boldsymbol{\Theta}}(x_i, \boldsymbol{\theta}, t) d\boldsymbol{\theta}}{\int_{\Omega_{\boldsymbol{\Theta}}} \tilde{p}_{X_i\boldsymbol{\Theta}}(x_i, \boldsymbol{\theta}, t) d\boldsymbol{\theta}} \tag{28}$$

Therefore, considering Eqs. (16), (19), and (28) we can write the equivalent decoupled form of the FPK equation,

$$\frac{\partial p_{X_i}(x_i,t)}{\partial t} = -\frac{\partial}{\partial x_i}\left(\tilde{f}_i(x_i,t)p_{X_i}(x_i,t)\right) + D_{ii}\frac{\partial^2 p_{X_i}(x_i,t)}{\partial x_i^2} \tag{29}$$

Recent studies have shown that the presented formulation can be generalized to a Globally-Evolving-Based General Density Evolution Equation (GE-GDEE) [29]. GE-GDEE is valid for any generic path-continuous stochastic process, whether or not it is Markovian [28].

White noise is an idealized or fictitious process with infinite power as no real physical process may have infinite signal power. Therefore, to use Eq. (29) in practice, one usually considers a band-limited white noise process that leads to some truncated version of the decomposed stochastic excitation. As it can be seen, an error in the approximation of the GDEE can directly affect the accuracy of results for the FPK equation. In this paper, the Stochastic Harmonic Function (SHF) of the second kind (SHF-II) [26] is used to approximate the stochastic excitation.

---

**Summary 1. Dimension-Reduced FPK Equation**

1. For a given appropriate stochastic differential equation, approximate the random excitation with a finite number of random variables via a decomposition scheme such as the Kosambi–Karhunen–Loève (KKL) or the Stochastic Harmonic Function (SHF) decompositions.
$$\dot{\mathbf{X}}_t = \mathbf{f}(t,\mathbf{X}_t) + \mathbf{g}(t,\mathbf{X}_t)\mathbf{h}(\boldsymbol{\Theta},t)$$

2. Solve the GDEE for each state and find $\tilde{p}_{X_i\boldsymbol{\Theta}}(x_i,\boldsymbol{\theta},t)$
$$\frac{\partial \tilde{p}_{X_i\boldsymbol{\Theta}}(x_i,\boldsymbol{\theta},t)}{\partial t} + \dot{X}_i(\boldsymbol{\theta},t)\frac{\partial \tilde{p}_{X_i\boldsymbol{\Theta}}(x_i,\boldsymbol{\theta},t)}{\partial x_i} = 0$$

3. Given $\tilde{p}_{X_i\boldsymbol{\Theta}}(x_i,\boldsymbol{\theta},t)$, compute the equivalent flux $\tilde{f}_i(x_i,t)$
$$\tilde{f}_i(x_i,t) = \frac{\int_{\Omega_{\boldsymbol{\Theta}}}\left(\dot{X}_i(\boldsymbol{\theta},t) - \sum_{j=1}^{m}\left(g_{ij}(t,X_i(\boldsymbol{\theta},t))h_j(\boldsymbol{\theta},t)\right)\right)\tilde{p}_{X_i\boldsymbol{\Theta}}(x_i,\boldsymbol{\theta},t)d\boldsymbol{\theta}}{\int_{\Omega_{\boldsymbol{\Theta}}}\tilde{p}_{X_i\boldsymbol{\Theta}}(x_i,\boldsymbol{\theta},t)d\boldsymbol{\theta}}$$

4. Solve the one-dimensional equivalent FPK equation
$$\frac{\partial p_{X_i}(x_i,t)}{\partial t} = -\frac{\partial}{\partial x_i}\left(\tilde{f}_i(x_i,t)p_{X_i}(x_i,t)\right) + D_{ii}\frac{\partial^2 p_{X_i}(x_i,t)}{\partial x_i^2}$$

---

## Network Setup

As universal function approximators [30], DNNs can be leveraged to establish complex and nonlinear mappings between input and output spaces. However, any DNN is essentially a finite composition of elementary operations. Therefore, one can obtain the analytical derivatives of a network utilizing the chain rule of differentiation. Automatic Differentiation (AD) uses this idea and can be leveraged to find the derivatives of a network either in the forward or the backward path without facing round-off or truncation errors [31]. The partial derivatives provided by AD can be used to encode differential equations within a DNN by defining the loss function as the differential equation in its residual form [18]. The DeepPDEM network [21] is a physics-informed DNN that utilizes this idea to encode the GDEE into the network. As stated earlier, to solve the equivalent decoupled formulation of the FPK equation, we first need to solve the GDEE. In this paper, we first introduce a modified concept of the DeepPDEM network. Later we use this modified version and the decoupled formulation to define the FPK-DP Net. To find the equivalent drift as required in Eq. (28), we need to integrate the spatial partial derivative of the density distribution over the probable space. This scheme is not present in Vanilla DeepPDEM. Therefore, in this research, we eliminate the marginalizing scheme from the DeepPDEM and handle the integrations explicitly.

Let $\tilde{p}_{X_i\Theta}(x_i, \theta, t)$ be a composition function incorporating deep networks $\mathcal{N}_G(x_i, \theta, t; \chi_G)$ that approximates the instantaneous solution to the GDEE, $p_{X_i}(x_i, \theta, t)$. The estimated $\tilde{p}_{X_i\Theta}(x_i, \theta, t)$ has to satisfy the GDEE and both the initial and the boundary conditions. The FPK equation is usually defined over an unbounded domain. Enforcing the boundary conditions for a numerical solver is usually challenging for an unbounded domain. In the proposed framework, we can either hard-code or soft-code the initial and boundary conditions into the network. The Initial Condition (IC) and the Boundary Condition (BC) hard-coding is achieved by including them explicitly in the composition function $\tilde{p}_{X_i\Theta}(x_i, \theta, t)$. On the other hand, penalizing the violation of the initial and boundary conditions in the loss function results in the soft-coded ICs/BCs. As confirmed in a previous study, though, hard-coding the boundary conditions offers superior performance [32]. Therefore, the IC/BC are hard-coded. Let $\wp_0(x, \boldsymbol{\theta})$ represent the initial condition. To hard-code the ICs and BCs into the network, the following functional form is assumed:

$$\tilde{\wp}_{X_i\boldsymbol{\Theta}}(x_i, \boldsymbol{\theta}, t) = \mathcal{N}_{G_1}(t) \cdot \exp\left(-\left(\sqrt{-\ln \wp_0(x_i, \boldsymbol{\theta})} + t\mathcal{N}_{G_2}(x_i, \boldsymbol{\theta}, t)\right)^2\right) \quad (30)$$

where $\mathcal{N}_{G_1}(\cdot)$ represents a network with strictly positive bounded output and $\mathcal{N}_{G_2}(\cdot)$ is a network with an unbounded output. From now on, we drop the known arguments to keep the formulations compact. A ReLU (Rectified Linear Unit – $max(0, x)$) activation function [33] is used for the output layer of $\mathcal{N}_{G_1}$ to ensure that the output remains positive. In addition, a scaled exponential activation function as given in Eq. (31) is utilized for the output layer of $\mathcal{N}_{G_2}$ to guarantee that its output is unbounded.

$$SEU(x) = c \, \text{sgn}(x)(\exp(|x|) - 1) \quad (31)$$

where $c$ is a hyper-parameter chosen to be equal to one for simplicity. The suggested structure ensures that the estimated PDF satisfies the initial condition and remains positive while evolving (1st axiom of the probability). $\mathcal{N}_{G_1}$ is also a scaling factor that allows the network to adjust the estimated density so that it remains normalized in accordance with the second axiom of the probability. At the end of each training epoch, the integral of the density is approximated numerically. This approximation is then used to update $\mathcal{N}_{G_1}$ accordingly so that the instantaneous density remains normal. thereby enforcing the 2nd axiom of the probability. Assuming that the output of $\mathcal{N}_{G_2}$ is unbounded and goes to infinity as the spatial inputs increase without bounds, the exponential term also satisfies the vanishing density boundary condition. For the partial derivatives of the approximator, we have:

$$\frac{\partial \tilde{p}_{X_i \Theta}}{\partial t} = \left( \frac{d\mathcal{N}_{G_1}}{dt} - 2\mathcal{N}_{G_1}(\sqrt{-\ln p_0} + t\mathcal{N}_{G_2})\left(\mathcal{N}_{G_2} + t\frac{\partial \mathcal{N}_{G_2}}{\partial t}\right) \right)$$
$$\cdot \exp\left(-(\sqrt{-\ln p_0} + t\mathcal{N}_{G_2})^2\right) \quad (32)$$

$$\frac{\partial \tilde{p}_{X_i \Theta}}{\partial x_i} = -2\mathcal{N}_{G_1}(\sqrt{-\ln p_0} + t\mathcal{N}_{G_2})\left( -\frac{\frac{dp_0}{dx}}{2p_0\sqrt{-\ln p_0}} + t\frac{\partial \mathcal{N}_{G_2}}{\partial x} \right)$$
$$\cdot \exp\left(-(\sqrt{-\ln p_0} + t\mathcal{N}_{G_2})^2\right) \quad (33)$$

To train a DNN and solve the density evolution, we have to convert the GDEE into an unconstrained optimization problem that is generally nonlinear and non-convex. To solve any GDEE of interest with the DeepPDEM concept, assume that the solution exists, is bounded, unique, and uniformly Lipchitz [21, 34]. Let "$\mathcal{L}_G$" be the approximation error (loss function) in $L^2$ sense. Since the boundary and initial conditions are already enforced as hard constraints, the loss function is used to enforce the GDEE in the residual form as a soft constraint.

$$\mathcal{L}_G(\tilde{p}_{X_i \Theta}) = \alpha_1 \left\| \frac{\partial \tilde{p}_{X_i \Theta}(x_i, \Theta, t; \chi_G)}{\partial t} + \dot{X}_i(\Theta, t) \frac{\partial \tilde{p}_{X_i \Theta}(x_i, \Theta, t; \chi_G)}{\partial x_i} \right\|_2^2, \quad \Theta \in \Omega_\Theta \quad (34)$$

where $\alpha_1$ is a strictly positive normalizing coefficient. Let "$\mathcal{L}_G^\star$" be the optimum loss. Therefore, the equivalent optimization problem is:

$$\mathcal{L}_G^\star(x_i, \Theta, t; \chi_G^\star) = \min_{\chi} \mathcal{L}_G(x_i, \Theta, t; \chi_G) \quad (35)$$

Here, we assume no prior knowledge of the instantaneous densities. If "$\mathcal{L}_G^\star$" is zero or sufficiently close to zero, the approximation "$\tilde{p}_{X_i \Theta}$" closely satisfies the GDEE and approaches the instantaneous PDF. To approximate Eq. (34), a finite set of collocation points can be used. Let $s_G$ be the number of collocation points chosen sufficiently large such that Eq. (36) represents the estimated loss with sufficient accuracy.

$$\hat{\mathcal{L}}_G(\tilde{p}_{X_i\boldsymbol{\Theta}}) = \frac{1}{s_G \hat{\mathcal{L}}_G^0} \sum_{j=1}^{s_G} \left\| \frac{\partial \tilde{p}_{X_i\boldsymbol{\Theta}}(x_i^j, \boldsymbol{\theta}^j, t^j)}{\partial t} + \dot{X}_i(\boldsymbol{\theta}^j, t^j) \frac{\partial \tilde{p}_{X_i\boldsymbol{\Theta}}(x_i^j, \boldsymbol{\theta}^j, t^j)}{\partial x_i} \right\|_2^2 \quad \boldsymbol{\theta} \in \Omega_{\boldsymbol{\theta}} \tag{36}$$

where the superscript shows the $j^{th}$ sample point, and $\hat{\mathcal{L}}_G^0$ represents the initial loss of the network. The SHF representation is used to decompose the stochastic excitation. Therefore, utilizing a space-filling Latin Hypercube Sampling (LHS) strategy [35], an i.i.d. sample set is generated in each training iteration to approximate the loss. Figure 1 depicts the DeepPDEM network training loop.

It is worth noting that DNNs usually demand a large quantity of labeled data to be trained that may not be available or be expensive to compile. However, the proposed method does not rely on any labeled training data, and the network learns the GDEE through randomly selected points from the domain. In essence, the DeepPDEM learns probability density evolution only by minimizing the GDEE in the residual form without using any labeled data.

Estimation of the marginal PDF $\tilde{p}_{X_i}(x_i, t)$ and subsequently the equivalent drift coefficient $\tilde{f}_i(x_i, t)$ is the next step. With a trained DeepPDEM surrogate of the GDEE at hand, for a given $x_i$ and $t$, the integration scheme shown in Figure 2 is used to integrate the probability density $\tilde{p}_{X_i\boldsymbol{\Theta}}(x_i, \boldsymbol{\theta}, t)$ over $\Omega_\Theta$.

The estimated equivalent drift coefficient can now be used to solve the decoupled equivalent form of the FPK equation (Eq. (29)). The DeepPDEM network scheme is summarized in Algorithm 1.

| Algorithm 1. DeepPDEM: Training the GDEE surrogate DNN |
|---|
| 5. Specify the training time interval $t \in [0, T]$ and the truncated spatial domain $D \subset \mathbb{R}$<br>6. Determine the SHF representation of the excitation $\mathbf{h}(\boldsymbol{\theta}, t)$, $\boldsymbol{\theta} \in \Omega_{\boldsymbol{\theta}}$<br>7. Compile the DeepPDEM network<br>8. Initialize the trainable variables $\boldsymbol{\chi}_G$ of the DeepPDEM network<br>9. Train the DeepPDEM network<br>   9.1. Sample the collocation points $\{t^j, x_i^j, \boldsymbol{\theta}^j\}_{j=1}^{s_G} \subset [0, T] \times D \times \Omega_{\boldsymbol{\theta}}$ |

| Algorithm 1. DeepPDEM: Training the GDEE surrogate DNN |
|---|
|     9.2. Compute the derivatives of the DeepPDEM network $\frac{\partial \tilde{p}_{X_i \Theta}}{\partial t}$ and $\frac{\partial \tilde{p}_{X_i \Theta}}{\partial x_i}$ <br>     9.3. Compute the loss $\hat{\mathcal{L}}_G(\tilde{p}_{X_i \Theta})$ <br>     9.4. Check for convergence <br>     9.5. Backpropagate and update the trainable variables $\chi_G$ <br> 10. Save the trained DeepPDEM surrogate $\chi_G^\star$ |

To solve Eq. (29), we use another physics-informed DNN called FPK-DP Net. Like the DeepPDEM network, once more assume that the solution to Eq. (29) exists, is bounded, unique, and uniformly Lipchitz. Now let $\hat{p}_{X_i}(x_i, t)$ be a composition function incorporating deep networks $\mathcal{N}_F(x_i, t; \chi_F)$ that approximates the instantaneous solution to Eq. (29), $p_{X_i}(x_i, t)$. A similar structure to the DeepPDEM is used for the FPK-DP Net.

$$\hat{p}_{X_i}(x_i, t) = \mathcal{N}_{F_1}(x_i, t) \cdot \exp\left(-\left(\sqrt{-\ln p_0(x_i)} + t\mathcal{N}_{F_2}(x_i, t)\right)^2\right) \quad (37)$$

Revisiting Eq. (29) leads to:

$$\frac{\partial p_{X_i}(x_i, t)}{\partial t} + \frac{\partial \tilde{f}_i(x_i, t)}{\partial x_i} p_{X_i}(x_i, t) + \frac{\partial p_{X_i}(x_i, t)}{\partial x_i} \tilde{f}_i(x_i, t) - D_{ii} \frac{\partial^2 p_{X_i}(x_i, t)}{\partial x_i^2} = 0 \quad (38)$$

Therefore, to form a similar loss function to Eq. (34), not only differentiation of the FPK-DP Net $\mathcal{N}_F(x_i, t; \chi_F)$ with respect to spatiotemporal variables is required, but also the derivatives of $\tilde{f}_i$ are needed as well. From Eq. (27) we have:

$$\frac{\partial \tilde{f}_i}{\partial x_i} = \frac{\frac{\partial \mu_i}{\partial x_i} \tilde{p}_{X_i} + \frac{\partial \tilde{p}_{X_i}}{\partial x_i} \mu_i}{(\tilde{p}_{X_i})^2} \quad (39)$$

By imposing the Leibniz integral rule assumptions over $\tilde{p}_{X_i \Theta}(x_i, \Theta, t)$, we have:

$$\frac{\partial \mu_i}{\partial x_i} = \int_{\Omega_\Theta} \left(\dot{X}_i(\Theta, t) - \sum_{j=1}^m \left(g_{ij}(t, X_i(\Theta, t)) h_j(\Theta, t)\right)\right) \frac{\partial \tilde{p}_{X_i \Theta}(x_i, \Theta, t)}{\partial x_i} d\Theta \quad (40)$$

$$\frac{\partial \tilde{p}_{X_i}}{\partial x_i} = \int_{\Omega_\Theta} \frac{\partial \tilde{p}_{X_i\Theta}(x_i, \theta, t)}{\partial x_i} d\theta \qquad (41)$$

The derivative $\frac{\partial \tilde{p}_{X_i\Theta}(x_i,\theta,t)}{\partial x_i}$ is already available from the DeepPDEM network (Figure 1). Thus, integrals (40) and (41) can be computed numerically giving the derivative $\tilde{f}_i$ with respect to the spatial variable using Eq. (39). For the second-order derivative of the PDF, we have:

$$\frac{\partial^2 \hat{p}_{X_i}}{\partial x_i^2} = \frac{\partial^2 \hat{p}_{X_i}(x_i, 0)}{\partial x_i^2} + t \cdot \eta \cdot \exp\left(-(\mathcal{N}_{G_2})^2\right) \qquad (42)$$

where $\eta$ is:

$$\eta = \frac{\partial^2 \mathcal{N}_{F_1}}{\partial x_i^2} - 2\mathcal{N}_{F_1}\mathcal{N}_{F_2}\frac{\partial^2 \mathcal{N}_{F_2}}{\partial x_i^2} + (4\mathcal{N}_{F_2}^2 - 2)\mathcal{N}_{F_1}\left(\frac{\partial \mathcal{N}_{F_2}}{\partial x_i}\right)^2 - 4\mathcal{N}_{F_2}\frac{\partial \mathcal{N}_{F_1}}{\partial x_i}\frac{\partial \mathcal{N}_{F_2}}{\partial x_i} \qquad (43)$$

All the derivatives in Eq. (43) are available through Automatic Differentiation (AD) of the networks $\mathcal{N}_{F_1}$ and $\mathcal{N}_{F_2}$. Now let "$\mathcal{L}_F$" be the approximation error (loss function) in $L^2$ sense defined as:

$$\mathcal{L}_F(\hat{p}_{X_i}) = \alpha_2 \left\|\frac{\partial \hat{p}_{X_i}}{\partial t} + \frac{\partial \tilde{f}_i}{\partial x_i}\hat{p}_{X_i} + \frac{\partial \hat{p}_{X_i}}{\partial x_i}\tilde{f}_i - D_{ii}\frac{\partial^2 \hat{p}_{X_i}}{\partial x_i^2}\right\|_2^2 \qquad (44)$$

where $\alpha_2$ is a strictly positive normalizing coefficient. Let $s_F$ represent the finite set of collocation points chosen sufficiently large such that Eq. (45) gives a sufficiently accurate estimation of the loss function, Eq. (44).

$$\hat{\mathcal{L}}_F(\hat{p}_{X_i}) = \frac{1}{s_F \hat{\mathcal{L}}_F^0} \sum_{j=1}^{s_F} \left\|\frac{\partial}{\partial t}\hat{p}_{X_i}(x_i^j, t^j) + \frac{\partial}{\partial x_i}\tilde{f}_i(x_i^j, t^j)\hat{p}_{X_i}(x_i^j, t^j) \right.$$
$$\left. + \frac{\partial}{\partial x_i}\hat{p}_{X_i}(x_i^j, t^j)\tilde{f}_i(x_i^j, t^j) - D_{ii}\frac{\partial^2}{\partial x_i^2}\hat{p}_{X_i}(x_i^j, t^j)\right\|_2^2 \qquad (45)$$

where the superscript indicates the sample point counter and $\hat{\mathcal{L}}_F^0$ is the initial loss of the network. In each training iteration, an i.i.d. collocation sample points set is generated using space-filling LHS [35]. The training loop is depicted in Figure 3 and Algorithm 2 outlines the FPK DP Net solution procedure.

---

Algorithm 2. FPK DP Net: Training a DNN to solve the FPK equation

---

1. Specify the training time interval $t \in [0, T]$ and the truncated spatial domain $D \subset \mathbb{R}^n$
2. Load the DeepPDEM network $\hat{p}_{X_i}$
3. Compile the FPK network
4. Initialize the trainable variables $\chi_F$ of the FPK network
5. Train the FPK network
   5.1. Sample the collocation points $\{t^j, x_i^j\}_{j=1}^{s_F} \subset [0, T] \times D$
   5.2. Compute DeepPDEM network $\hat{p}_{X_i}$ outputs for the sample points
      5.2.1. Compute the equivalent drift coefficient $\tilde{f}_i(x_i^j, t^j)$ and the derivative $\frac{\partial}{\partial x_i} \tilde{f}_i(x_i^j, t^j) \hat{p}_{X_i}$
   5.3. Compute the derivatives of the FPK network $\frac{\partial}{\partial t} \hat{p}_{X_i}(x_i^j, t^j)$ and $\frac{\partial}{\partial x_i} \hat{p}_{X_i}(x_i^j, t^j)$
   5.4. Compute the loss $\hat{\mathcal{L}}_F(\hat{p}_{X_i})$
   5.5. Check for convergence
   10.1. Backpropagate and update the trainable variables $\chi_F$
6. Save the trained FPK DP network $\chi_F^\star$

---

Even though Physics-Informed Neural Networks may sometimes fail in the training process [36-38], it has been shown that Multi-Layer Perceptron (MLP) is sufficient for dealing with classical PDEs encountered in solid, fluid, and quantum mechanics, [18, 39, 40]. Although an MLP network with a sufficient number of neurons can approximate a function and its partial derivatives [41], determining a sufficient number of neurons is not always straightforward. It may change from one problem to another. In this study, we used networks with a depth of 4 layers, and a width of 20 neurons for each network explained earlier. Increasing the depth of the network increases the computational complexity and thus decreases the convergence rate significantly. To choose the network dimensions, via trial and error, assurances are made that the FPK-DP Net converges to an admissible solution for all test cases. To this aim, the networks are trained for the chosen dimension. If the optimization process reduces the loss function below $1E - 3$ limit for all test cases, the choice is considered admissible. In addition, for each test case, the results are compared with those of the Monte-Carlo simulations that showed good compatibility. In other words, similar consistent behavior was realized as long as the loss values were close to 1E-3. Through our experiments, we concluded that the network depth has a significant role in overfitting and/or premature convergence.

The Glorot normal initializer [42] was utilized in our networks to initialize the biases and the weights. Traditional activation functions, such as the hyperbolic tangent, can suffer from vanishing gradients. Swish is a non-monotonic, smooth function that helps to alleviate this problem. This can lead to improved convergence rates for neural networks. Thus, the Swish activation function is used for the hidden layers [21].

Given the large number of trainable variables in a deep network, a gradient-based optimizer is the only viable choice. Although, utilizing gradient-based optimizers the chance of converging to a local (and sometimes bad) optima. Employing an adaptive scheme for the learning rate is one of the possible strategies to avoid premature convergence in stochastic gradient descent algorithms such as Adam [43, 44]. By keeping the learning rate small for the first few epochs of training and then increasing it gradually, we can achieve a good balance between exploration and exploitation in the solution space. The training rate plays an important role in the training process and can affect the training and convergence rate, significantly [45]. In this respect, the rectified adaptive learning rate [44] was used to update the learning rate in the training process. Technically, as explained earlier, we would like the learning rate to be as large as possible initially and gradually decrease in the process. Therefore, we chose the highest value to start from in the adaptive learning rate algorithm such that none of our networks diverge, i.e., 0.015.

## Numerical Study and Analysis

The proposed method can be used to solve the FPK equation with no restrictions on the linearity or dimension of the stochastic dynamic. To demonstrate the applicability of the proposed method, five case studies are discussed in this section. First, two one-dimensional cases in which the equivalent drift is known and readily available, are considered. These two test cases help validate the proposed method. Subsequently, we consider two different two-dimensional nonlinear dynamics, namely the Duffing and the FitzHugh-Nagumo oscillators. Finally, an investigation of a three-dimensional Roessler oscillator concludes our numerical studies. FPK-DP Net does not have any restrictions on dimensionality and theoretically, it should be able to handle higher-dimensional problems. However, the test cases are selected as such to minimize the complications and better analyze and discuss the behavior of the network. Monte-Carlo simulations are used to verify the results, which is challenging for higher dimension problems.

**Case-Study I**

For the first case study, we may consider a one-dimensional SDE,

$$dX = (\alpha X - \beta X^3)dt + \sigma dB \tag{46}$$

where $\alpha$ and $\beta$ are real positive values, and $\sigma$ is the noise intensity. Eq. (47) shows the FPK equation for this SDE.

$$\frac{\partial}{\partial t}p_X(x) = \frac{\sigma^2}{2}\frac{\partial^2}{\partial x^2}p_X(x) + \frac{\partial}{\partial x}\left((\beta x^3 - \alpha x)p_X(x)\right) \tag{47}$$

We may assume $\beta = \sigma = 0.5$ and a standard normal distribution for the initial condition $X_0 \sim N(0,1)$. The exact steady-state solution for this equation is readily available as presented in Eq. (48).

$$p_{X_{ss}}(x) = \frac{\exp(4\alpha x^2 - x^4)}{K_{\frac{1}{4}}(2\alpha^2)\sqrt{\alpha}\exp(-2\alpha^2)} \tag{48}$$

Given the steady-state solution, a well-informed estimate can be used to truncate the solution space. To solve the GDEE, the stochastic excitation is approximated using a Stochastic Harmonic Function of the second kind with 25 random parameters $\Theta_i$. Figure 4 (a) depicts a realization of random parameters $\theta_i$. Figure 4 (b) shows the Power Spectral Density (PSD) of the associated Stochastic Harmonic Function (SHF) estimation.

By choosing three different values for $\alpha$, we can benchmark the capability of the proposed method in solving the problem at hand. We use the Monte-Carlo simulation with $10^5$ samples to compare the results. Two snapshots of time, at $t = 0.2$ and $1.8$, and three different values for the parameter $\alpha$ were chosen for the comparison. We trained the FPK-DP network for a two-second time interval. Figure 5 shows the predicted values by the FPK-DP Net after $10^6$ iterations for three different values of $\alpha$. Monte-Carlo simulation results are also shown in bar charts for each case, representing the relative frequency of occurrence. For each case the loss value $\hat{\mathcal{L}}_F$ suggests a good network convergence. Comparing the results with the Monte-Carlo simulation also shows satisfactory agreement. Table 1 gives the relative Entropy of the obtained probability distributions with respect to the steady-state and the standard normal distributions.

Table 1 Relative Entropy (Kullback-Leibler divergence) of the solution with respect to steady-state response and standard normal probability distribution

|  | Time | Steady-State | Standard Normal |
|---|---|---|---|
| $\alpha = 0.3$ | 0.2 | $-0.1174$ | 0.0558 |
|  | 1.6 | $-0.0091$ | 0.1044 |
| $\alpha = 0.5$ | 0.2 | $-0.1701$ | 0.0293 |
|  | 1.6 | $-0.0498$ | 0.1068 |
| $\alpha = 0.7$ | 0.2 | $-0.3844$ | 0.0089 |
|  | 1.6 | $-0.0914$ | $-0.1121$ |

Figure 6 represents the estimated PDF evolution by the FPK-DP Net for the case $\alpha = 0.3$ and $t \in [0,2]$ which, as expected, shows a smooth transition from the initial condition to the steady-state solution.

A good checkpoint for this problem is to analyze the estimated equivalent drift by the FPK-DP network. As depicted in Figure 7 and represented in Table 2, the estimated equivalent drift coefficient follows the analytical solution with an acceptable error; however, it is evident that as the time increases, so does the Root Mean Square Error (RMSE) of the estimated equivalent drift coefficient. Although this does not affect the final results, it could be anticipated since the dimension reduction scheme approximates the stochastic input. This increase in RMSE could lead to an underestimation of the equivalent drift coefficient.

Table 2 Comparison of the FPK-DP Net equivalent drift estimation error

| Time | Root Mean Square Error (RMSE) | | |
| --- | --- | --- | --- |
| | $\alpha = 0.3$ | $\alpha = 0.5$ | $\alpha = 0.7$ |
| 0.2 | 0.0187 | 0.0449 | 0.0259 |
| 0.4 | 0.0481 | 0.0467 | 0.0511 |
| 0.8 | 0.0421 | 0.0688 | 0.0555 |
| 1.6 | 0.0505 | 0.0475 | 0.0574 |

We may next analyze the outputs of the nested networks $N_{F_1}(t)$ and $N_{F_2}(x,t)$ to check the inner working of the FPK-DP network. As explained earlier, we designed the $N_{F_1}(t)$ network such that its output is strictly positive. Furthermore, since the output of the FPK-DP network is forced to start from the initial condition, it is anticipated that the output of the $N_{F_1}(t)$ network should start from one. As the $N_{F_1}(t)$ network output evolves through time, it regulates and satisfies the nonlocal integral constraint. Figure 8 depicts the output of the $N_{F_1}(t)$ network for the three chosen values of the parameter $\alpha$. The network's output starts from the unity, as anticipated, and decreases over time. Two outputs show strictly monotonic behavior in the time interval of interest. However, this monotonicity in the behavior is not a general characteristic of the $N_{F_1}(t)$ output. On the other hand, as the solution presents, the outputs for all the parameter values drop below one which could indicate that after a while $N_{F_2}(t,x)$ starts to overestimate the density. A closer look at the equivalent drift coefficient and $N_{F_2}(t,x)$ outputs, Figure 9, suggests that this overestimation primarily happens at the tail values of the probability density distribution.

Opposite to $N_{F_1}(t)$, the $N_{F_2}(t)$ network is designed to increase its output as the spatial variable increases without bounds. Figure 9 represents the output of the $N_{F_2}(t,x)$ network for the three chosen values of the parameter $\alpha$. As expected, all the outputs tend to increase without bounds as the spatial distance increases from the origin.

**Case-Study II**

For the second test case, consider the nonlinear SDE given in Eq. (49). Once again, $\sigma$ is the noise intensity, and $\alpha$ is a parameter chosen equal to 0.5.

$$dX = \left(\alpha - X + \frac{X^8}{1 + X^8}\right)dt + \sigma dB \tag{49}$$

The FPK equation for the SDE Eq. (49) is given in Eq. (50). The exact steady-state solution for Eq. (50) is also readily available.

$$\frac{\partial}{\partial t}\wp_X(x) = \frac{\sigma^2}{2}\frac{\partial^2}{\partial x^2}\wp_X(x) + \frac{\partial}{\partial x}\left(\left(\alpha - X + \frac{X^8}{1 + X^8}\right)\wp_X(x)\right) \tag{50}$$

A standard normal distribution for the initial condition $X_0 \sim N(0,1)$ is assumed. Given the steady-state solution, the solution space can be truncated for the proposed numerical scheme. Like the previous test case, a Stochastic Harmonic Function of the second kind with 25 random parameters $\Theta_i$ are used to approximate the stochastic excitation and solve the GDEE. Once again, a $10^5$ sample Monte-Carlo simulation, as well as four snapshots of time, at $t = 0.95, 4.75, 9.5,$ and $19$, were chosen for comparison (Figure 10). The FPK-DP network was trained for $2 \times 10^7$ iterations to solve the twenty seconds time interval. The converged loss value $\hat{\mathcal{L}}_F = 1.8338E - 6$ is relatively higher than the previous test case. This increase can be justified, given the increase in the temporal dimension. On the other hand, the PDF estimated by the network shows good consistency with the Monte-Carlo simulation results. Table 3 gives the relative Entropy of the obtained probability distributions with respect to the steady-state and the standard normal distributions.

Table 3 Relative Entropy (Kullback-Leibler divergence) of the second test case solution with respect to its steady-state response and standard normal probability distribution

| Time | Steady-State | Standard Normal |
|------|--------------|-----------------|
| 0.95 | −0.2028 | 0.2477 |
| 4.75 | −0.0265 | 0.1078 |
| 9.5  | −0.0191 | 0.0442 |
| 19   | −0.0173 | 0.0240 |

Figure 11 depicts the calculated PDF evolution by the FPK-DP Net for $t \in [0,20]$ which, represents the transition from the initial condition to the steady-state solution. The output of the $N_{F_1}(t)$ network, in this case, remained strictly positive and relatively close to one, as shown in Figure 12 (a). As anticipated earlier, the monotonicity in the behavior $N_{F_1}(t)$ was not seen. Figure 12 (b) represents the outputs of the $N_{F_2}(t,x)$ network in which the output convergence to a steady-state value is evident.

**Case-Study III**

For the third test case, we expand our study and consider another classical nonlinear oscillator. The FitzHugh–Nagumo oscillator, Eq. (51), is a relaxation oscillator commonly used in electronics.

$$d\mathbf{X} = \begin{bmatrix} X_1 - \frac{X_1^3}{3} - X_2 + I_0 \\ C_1 \cdot (X_1 + C_2 - C_3 X_2) \end{bmatrix} dt + \begin{bmatrix} \sigma \\ 0 \end{bmatrix} dB \tag{51}$$

The FPK equation for the FitzHugh–Nagumo oscillator is given in Eq. (52).

$$\frac{\partial}{\partial t}p_{\mathbf{X}}(\mathbf{x}) = \frac{\sigma^2}{2}\frac{\partial^2}{\partial x_1^2}p_{\mathbf{X}}(\mathbf{x}) + \frac{\partial}{\partial x_1}\left(\left(x_1 - \frac{x_1^3}{3} - x_2 + I_0\right)p_{\mathbf{X}}(\mathbf{x})\right) \\ + \frac{\partial}{\partial x_2}\left((C_1 \cdot (x_1 + C_2 - C_3 x_2))p_{\mathbf{X}}(\mathbf{x})\right) \tag{52}$$

Assuming $I_0 = -1, C_1 = 1, C_2 = 0.6, C_3 = 0.25, \sigma = 0.6$, the FPK-DP Net estimated solution for the first state, $X_1$, is shown in Figure 13. We also assume a Gaussian distribution with a tighter standard deviation for the initial condition with 0.1 mean and 0.01 standard deviation, $\mathbf{X_0} \sim N(0.1, 0.01)$. A $10^5$ sample Monte-Carlo simulation, and four snapshots of time, at $t = 0.95, 1.25, 3.5,$ and $4.75$ were chosen for the comparison. The FPK-DP network was trained for $1 \times 10^6$ iterations to solve the five seconds time interval. The converged loss value is $\hat{\mathcal{L}}_F = 7.9639E-5$.

Since a tighter standard deviation is used for the initial conditions, it is expected that the nonlinearity in the output of the $N_{F_2}(t,x)$ network should be shifted towards larger values of the spatial variable. The network behaves more linearly compared with the previous test cases. Figure 14 gives the $N_{F_2}(t,x)$ outputs for the chosen snapshots of time. As expected, over the domain of interest, the network provided a semi-linear output which also shows a consistent behavior with an increase of the standard deviation over time.

**Case-Study IV**

Within the previous test cases, we studied the performance of the FPK-DP Net for two autonomous dynamics. In this test case, the performance of the proposed network for a non-autonomous dynamic is analyzed. The Duffing oscillator given in Eq. (53), is chosen as the next test case:

$$\ddot{x} + \eta \dot{x} + \alpha x + \beta x^3 = \gamma \cos(\omega t) \tag{53}$$

Considering an additive Brownian excitation, Eq. (53) leads to the following SDE:

$$d\mathbf{X} = \begin{bmatrix} X_2 \\ -\eta X_2 - \alpha X_1 - \beta X_1^3 + \gamma \cos(\omega t) \end{bmatrix} dt + \begin{bmatrix} 0 \\ \sigma \end{bmatrix} dB \tag{54}$$

We first consider the autonomous formulation, $\gamma = 0$, since the exact steady-state solution of the FPK equation, given in Eq. (55), is known (Eq. (56)). Then, we move forward to the more evolved non-autonomous dynamic where $\gamma \neq 0$.

$$\frac{\partial}{\partial t}p_{\mathbf{X}}(\mathbf{x}) = \frac{\sigma^2}{2}\frac{\partial^2}{\partial x_2^2}p_{\mathbf{X}}(\mathbf{x}) + \frac{\partial}{\partial x_1}(x_2 p_{\mathbf{X}}(\mathbf{x}))$$
$$+ \frac{\partial}{\partial x_2}\left((-\eta x_2 - \alpha x_1 - \beta x_1^3 + \gamma \cos(\omega t))p_X(x)\right) \tag{55}$$

$$p_{\mathbf{X}_{ss}}(\mathbf{x}) = C(\eta,\alpha,\beta) \cdot \exp\left(-\frac{\eta}{\sigma^2}\left(x_2^2 + \alpha x_1^2 + \frac{\beta}{2}x_1^4\right)\right) \tag{56}$$

In Eq. (55), $C(\eta,\alpha,\beta)$ is the normalization constant. Let $\alpha = -1$, $\beta = 1$, $\eta = 0.3$ and $\sigma = 0.57$. We also assumed a Gaussian distribution for the initial condition with 0.1 mean and 0.01 standard deviation, $\mathbf{X_0} \sim N(0.1, 0.01)$. Once again, a $10^5$ sample Monte-Carlo simulation, and four snapshots of time, at $t = 0.95, 1.25, 3.5,$ and $4.75$ were chosen for the comparison. The FPK-DP network was trained for $1 \times 10^6$ iterations. The converged loss value is $\hat{\mathcal{L}}_F = 2.3121E - 5$. Figure 15 depicts the results obtained for the second state, $X_2$. Compared with the Monte-Carlo simulation, the FPK-DP Net provided a consistent solution for the FPK equation. Table 4 also gives the Kullback-Liebler divergence of the solution with respect to steady-state and standard normal probability density distribution.

Table 4 Relative Entropy of the obtained results with respect to steady-state and standard normal distributions

| Time | Relative Entropy (with respect to) | |
|---|---|---|
| | Steady-State Response | Standard Normal Distribution |
| 0.95 | 0.3300 | 0.3483 |
| 1.25 | 0.0438 | 0.0590 |
| 3.5 | −0.0163 | −0.0675 |
| 4.75 | 0.0129 | 0.0241 |

Another figure of merit that can be addressed for this case is the PDF tail behavior. Figure 16 depicts the evolution of the estimated solution on the logarithmic scale. For all snapshots of time, the estimated behavior is strictly decreasing, which is in accordance with the expected behavior of a PDF.

Now, let $\gamma = 0.65$ and $\omega = 1.2$. To estimate the solution for this non-autonomous case, we used the weights found in the autonomous form to initialize the FPK-DP Net and trained the network for $1 \times 10^4$ iterations. The loss value is $\hat{\mathcal{L}}_F = 8.3332E - 5$, and the comparison of the results is shown in Figure 17.

The behavior of the tail of the PDF is no longer strictly decreasing, as seen in Figure 18. Although, the results still suggest a consistent solution.

Figure 19 compares the outputs of the $N_{F_2}(t, x)$ network for the autonomous and the non-autonomous Duffing oscillator in the chosen snapshots of time. As expected, the network represents a semi-linear behavior.

As the results suggest, FPK-DP Net provides a consistent estimation of the PDF. On the other hand, given the initialization policy, the network trained relatively faster. The results suggest that the proposed initialization strategy could be leveraged for complicated systems in which the network can be trained iteratively with approximations of the system. However, testing this idea requires further studies that are beyond the purpose and scope of the current work.

**Case-Study V**

The last test case studied in this paper is the Roessler attractor, Eq. (57), which is a three-dimensional self-sustained oscillator.

$$d\mathbf{X} = \begin{bmatrix} -X_2 - X_3 \\ X_1 + C_1 X_2 \\ C_2 + X_3(X_1 - C_3) \end{bmatrix} dt + \begin{bmatrix} 0 \\ 0 \\ \sigma \end{bmatrix} dB \qquad (57)$$

The FPK equation for the Roessler oscillator is given in Eq.(58).

$$\frac{\partial}{\partial t}p_{\mathbf{X}}(\mathbf{x}) = \frac{\sigma^2}{2}\frac{\partial^2}{\partial x_1^2}p_{\mathbf{X}}(\mathbf{x}) + \frac{\partial}{\partial x_1}\left((-x_2 - x_3)p_{\mathbf{X}}(\mathbf{x})\right)$$

$$+ \frac{\partial}{\partial x_2}\left((x_1 + C_1 x_2)p_{\mathbf{X}}(\mathbf{x})\right) \quad (58)$$

$$+ \frac{\partial}{\partial x_3}\left((C_2 + x_3 \cdot (x_1 - c_3))p_{\mathbf{X}}(\mathbf{x})\right)$$

Let $C_1 = 0.2$, $C_2 = 2$, $C_3 = 5.7$ and $\sigma = 0.5$. Like the previous test case, Gaussian distributions were assumed for the initial conditions with 0.1 mean and 0.01 standard deviation, $\mathbf{X_0} \sim N(0.1, 0.01)$. The FPK-DP network was trained for $1 \times 10^6$ iterations to solve the five seconds time interval. The converged loss value is $\hat{\mathcal{L}}_F = 3.99456E - 4$. A Monte-Carlo simulation with $10^5$ samples and four snapshots of time, at $t = 0.95, 1.25, 3.5,$ and $4.75$, were chosen for the comparison. The solution found by utilizing the FPK-DP Net is given in Figure 20 and Figure 21. Table 5 also provides the moments of the obtained probability density.

Table 5 Moments of the obtained solution

| Time | Mean | Variance | Skewness | Kurtosis |
|---|---|---|---|---|
| 0.95 | 0.0504 | 0.0024 | −0.0252 | 3.1855 |
| 1.25 | −0.0498 | 0.0041 | −0.0267 | 3.1405 |
| 3.5 | −1.0063 | 0.0161 | 0.0341 | 2.9877 |
| 4.75 | −0.5650 | 0.0280 | 0.0510 | 3.0283 |

## Concluding Remarks

The evolution of the probability density for stochastical dynamic systems is of practical significance for many engineering applications. To this aim, the Fokker-Plank-Kolmogorov (FPK) equation, as an idealized model that quantifies and captures the evolution of the probability density is solved using a novel machine learning approach. However, since a generic solution for high dimensional complex and coupled FPK equation is still an open problem, the FPK-DP Net as a novel mesh-free physics-informed DNN framework is proposed in the present study. In addition, a decoupling scheme is also presented that allows for easier handling of the training process via breaking the main problem into several smaller and more straightforward sub-problems. The decoupling scheme will also help break down high-dimensional FPK equations into several one-dimensional equivalent FPK equations. The proposed FPK-DP Net framework is verified for several relevant test cases using a Monte Carlo approach. Comparisons of the results demonstrate the efficacy of the proposed FPK-DP Net for solving the FPK equation with various initial and boundary conditions. Another vital advantage of the new approach is that, unlike many other DNNs, the FPK-DP Net does not require any labeled data for training. Moreover, as the FPK-DP Net utilizes a dimension-reduced formulation, it can handle high-dimensional practical problems as well. Besides, FPK-DP Net can also be used as an efficient surrogate model for controller design, as well as design optimization of any stochastic system.

## Declarations and Compliance with Ethical Standards

### Conflict of interest

The authors declare that they have no conflict of interest.

### Funding

This study was not funded.

### Data Availability

The datasets generated and/or analyzed during the current study are available from the authors upon request.

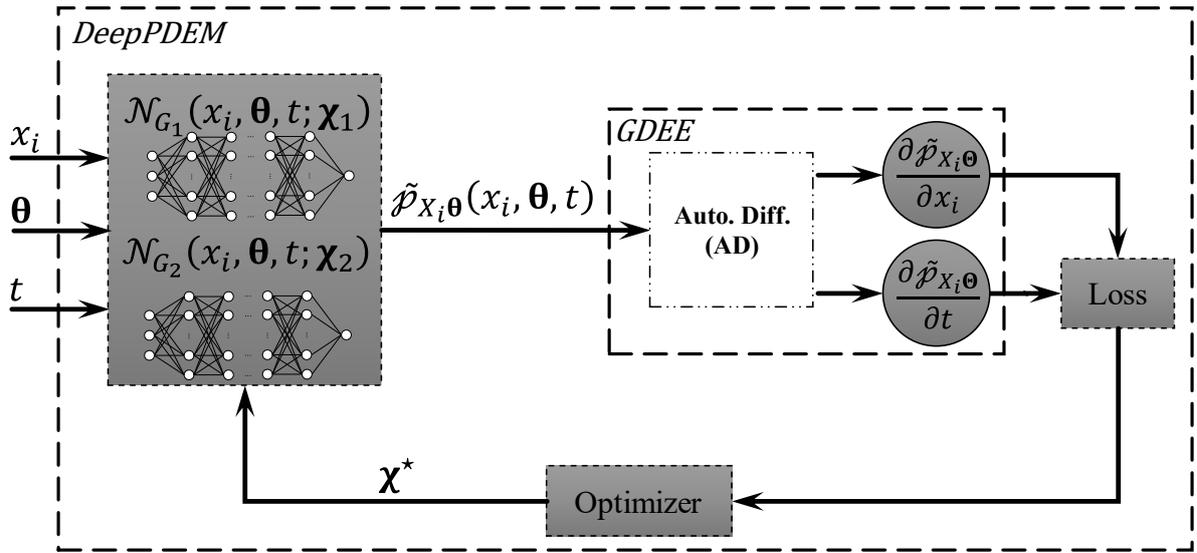

Figure 1 DeepPDEM Network

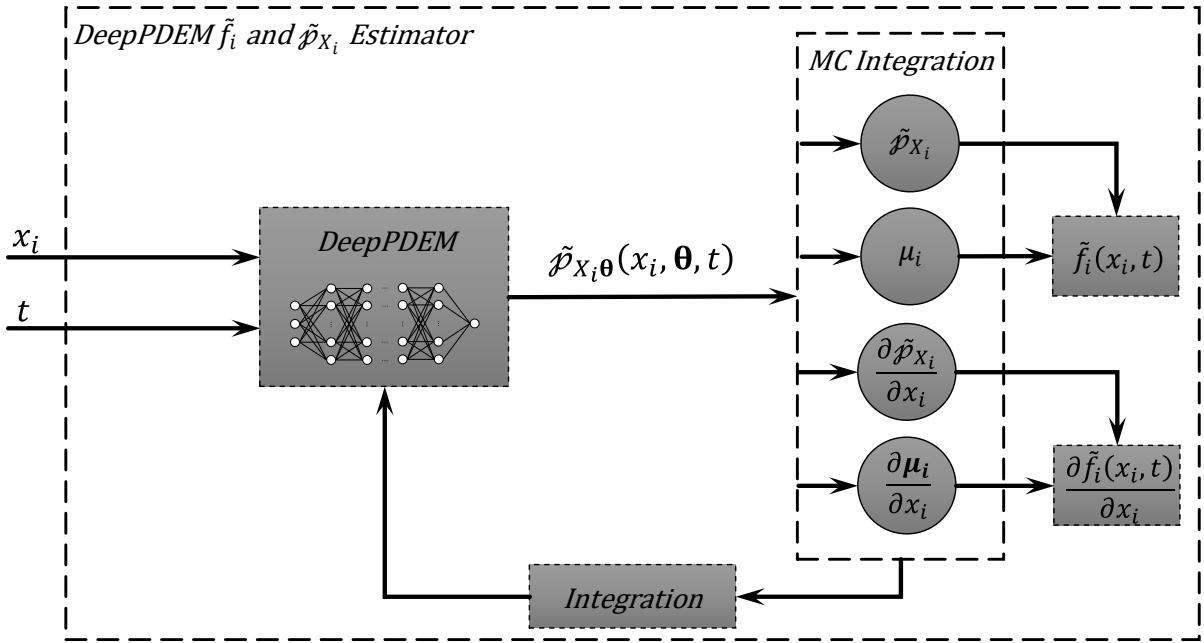

Figure 2 Marginal PDF and equivalent drift coefficient estimation

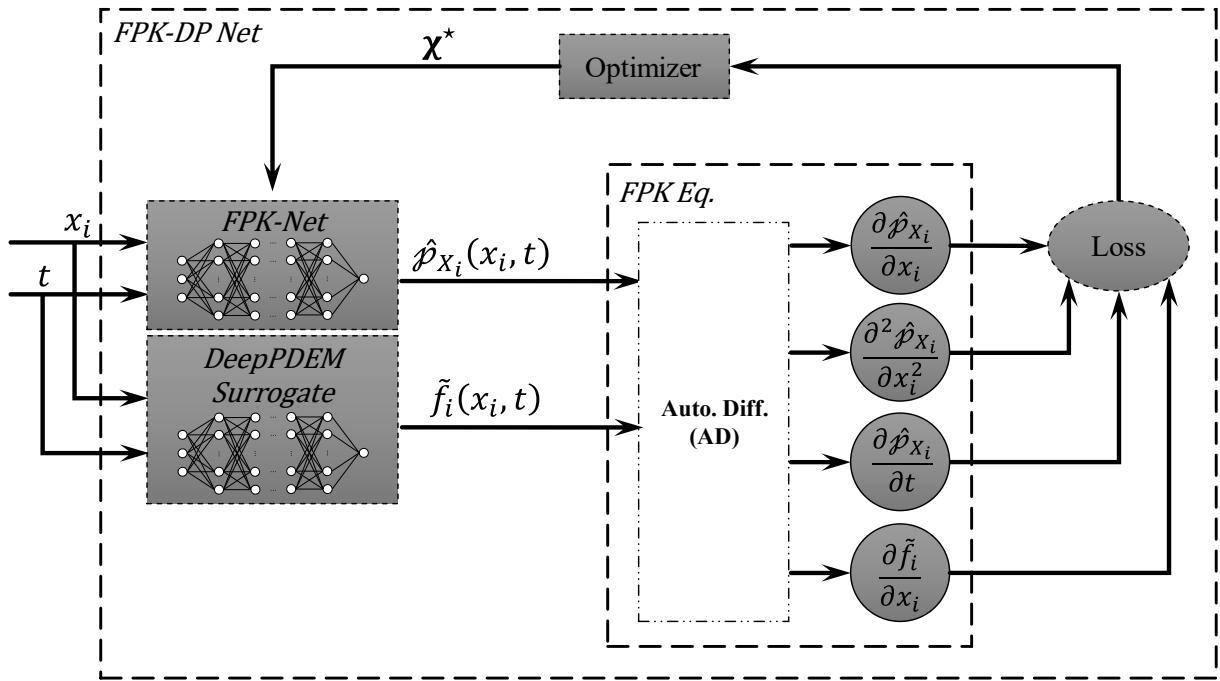

Figure 3 The DNN scheme to solve the FPK equation

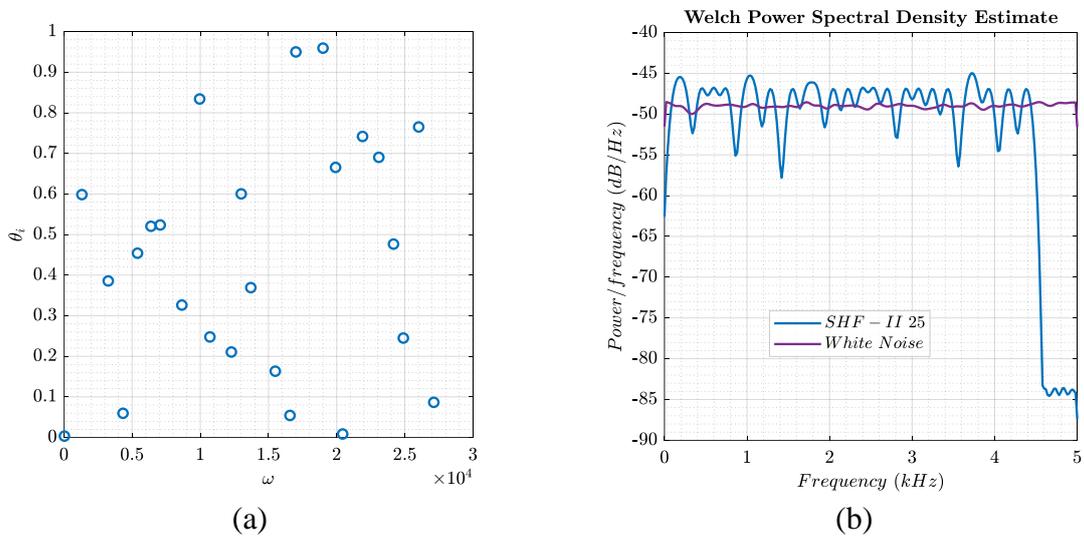

(a)   (b)

Figure 4 A realization of SHF-II approximation of the stochastic excitation

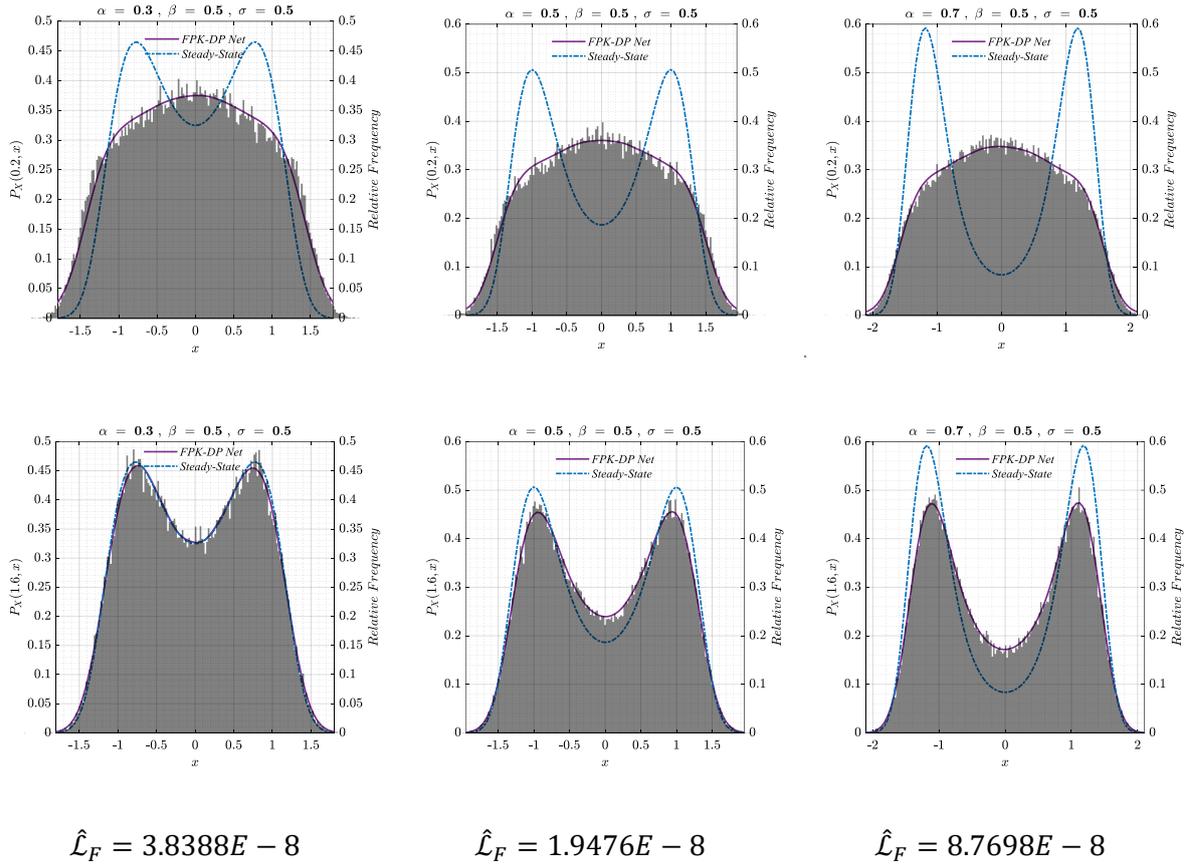

$$\hat{\mathcal{L}}_F = 3.8388E-8 \qquad \hat{\mathcal{L}}_F = 1.9476E-8 \qquad \hat{\mathcal{L}}_F = 8.7698E-8$$

Figure 5 – Comparison of the FPK-DP Net predicted PDFs with Monte-Carlo simulation results with different values of $\alpha$

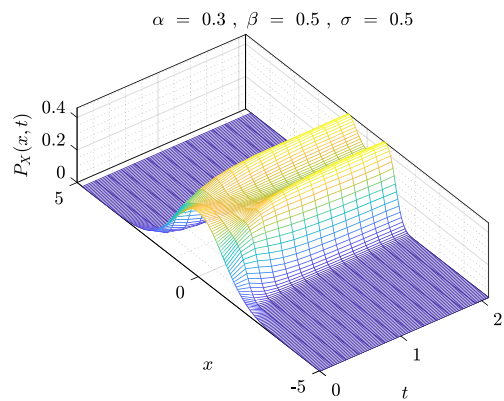

Figure 6 – The evolution of PDF predicted by the FPK-DP Net where $\alpha = 0.5$ and the initial condition is standard normal

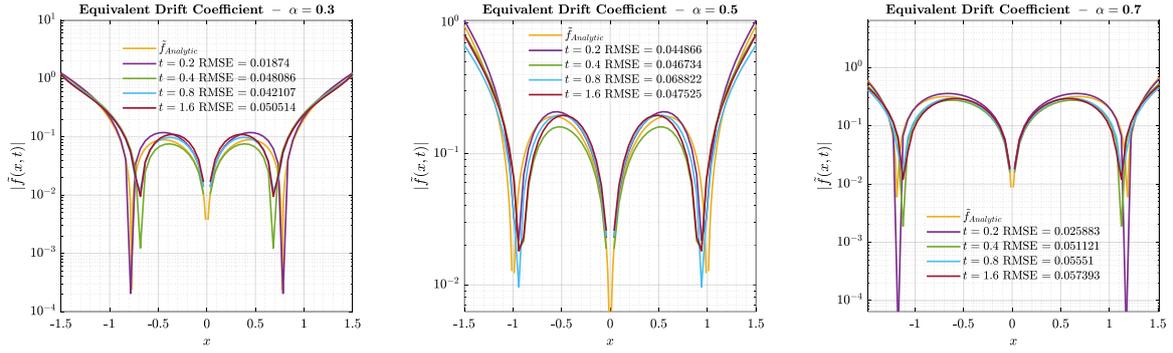

Figure 7 – Comparison of the FPK-DP Net estimated equivalent drift coefficient for different values of $\alpha$

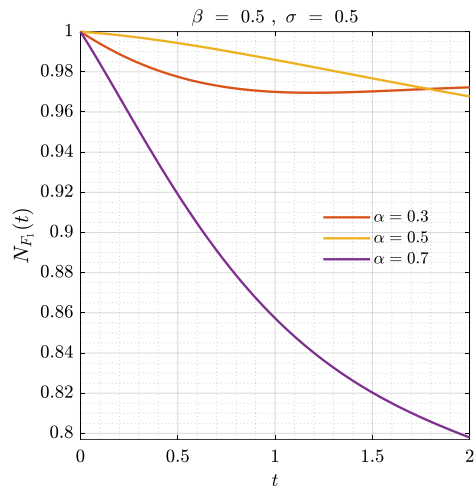

Figure 8 - Comparison of the $N_{F_1}(t)$ network outputs for different values of the parameter $\alpha$

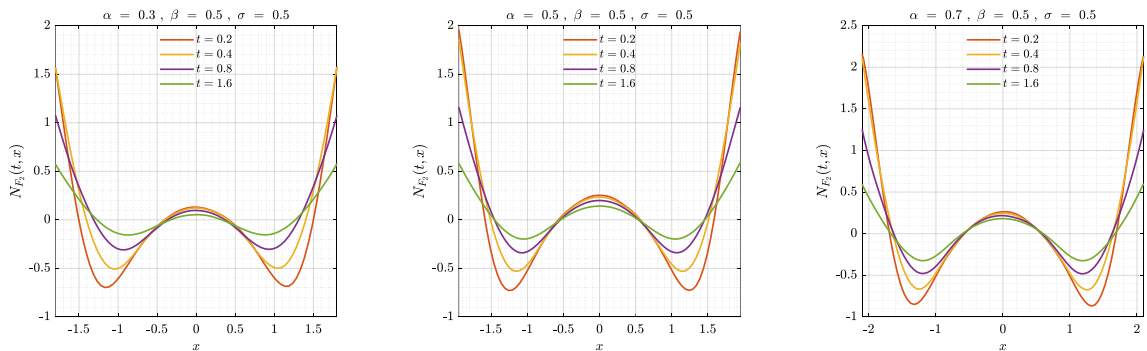

Figure 9 – Comparison of the $N_{F_2}$ network output for the three values of the parameter $\alpha$ in four snapshots of

time

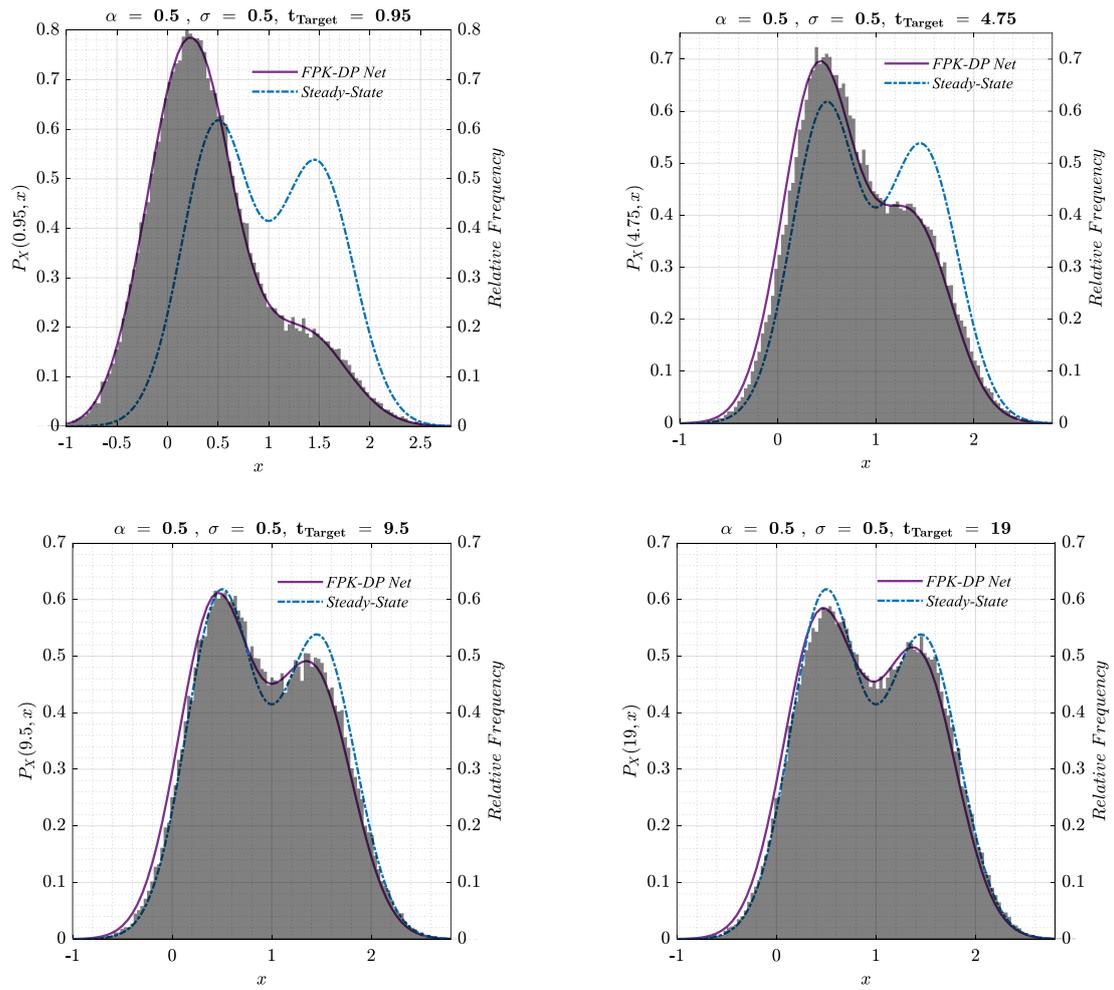

$\hat{\mathcal{L}}_F = 1.8338E - 6$

Figure 10 – Comparison of the FPK-DP Net predicted PDFs with Monte-Carlo simulation results for $\alpha = 0.5$

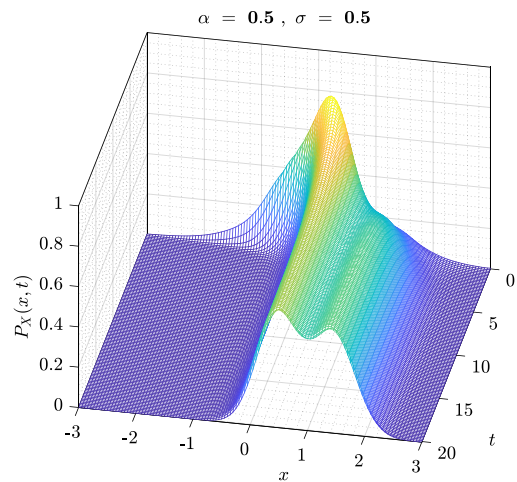

Figure 11 – The evolution of PDF predicted by the FPK-DP Net

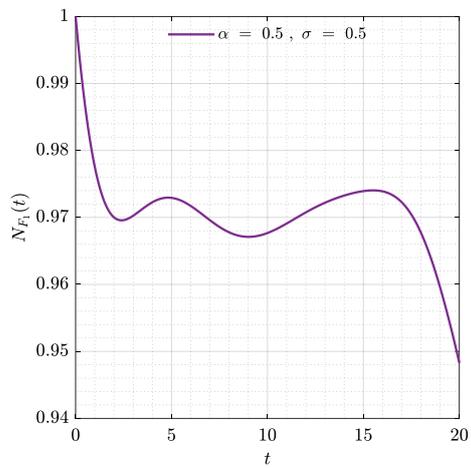
(a)

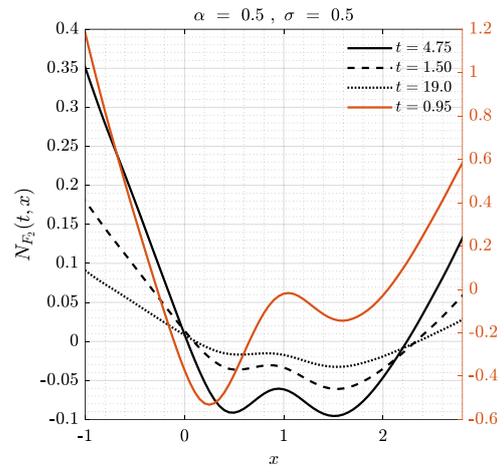
(b)

Figure 12 – The evolution of the $N_{F_1}$ network output (a) and the $N_{F_2}$ network output (b) over time

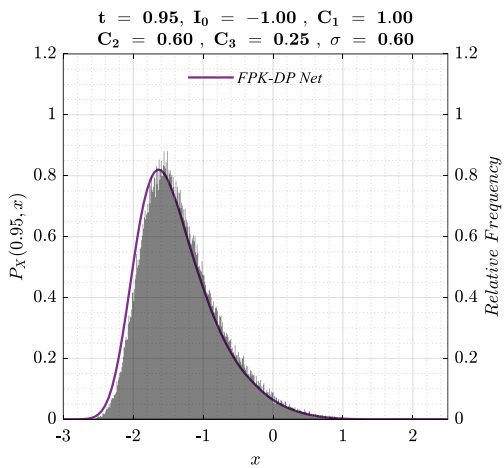

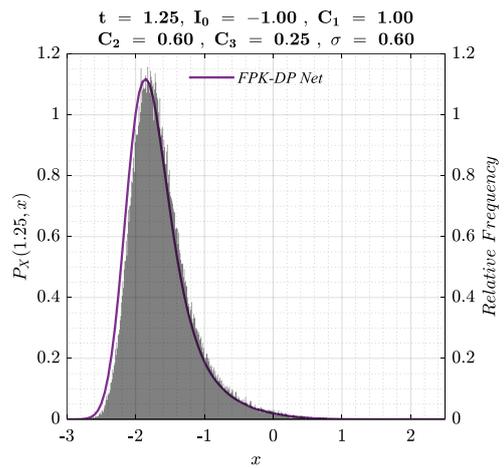

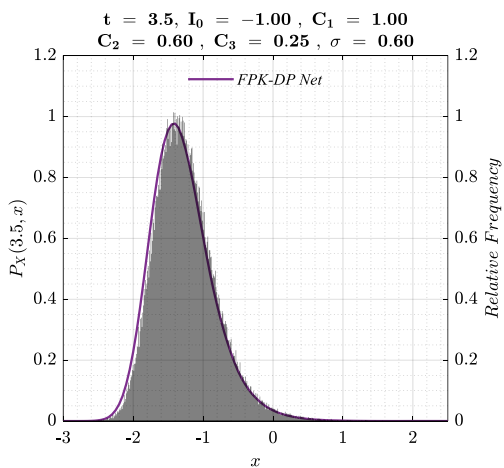

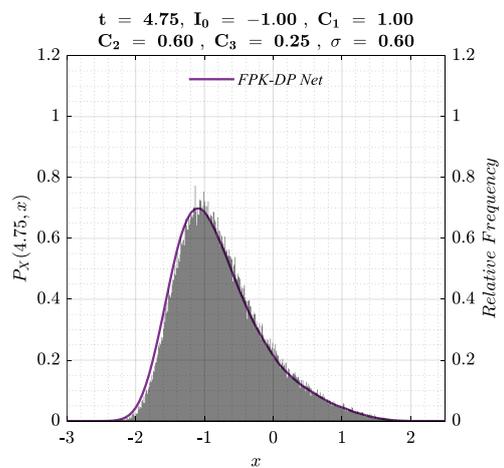

$$\hat{\mathcal{L}}_F = 7.96389E-5$$

Figure 13 – Comparison of the FPK-DP Net predicted PDF with Monte-Carlo simulation results for FitzHugh–Nagumo oscillator, where $I_0 = -1$, $C_1 = 1$, $C_2 = 0.6$, $C_3 = 0.25$, and $\sigma = 0.6$

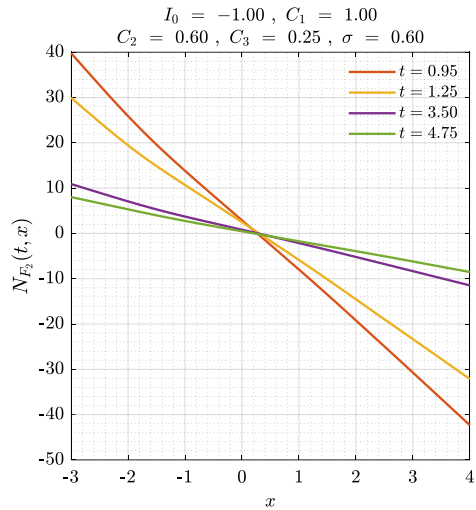

Figure 14 – Evolution of the output of the $N_{F_2}(t,x)$ network for the FitzHugh–Nagumo oscillator

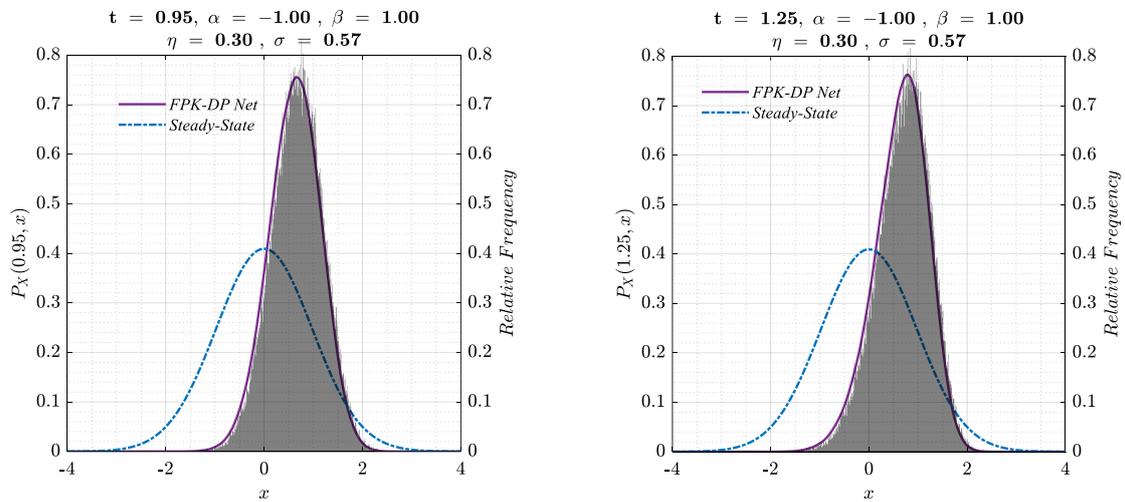

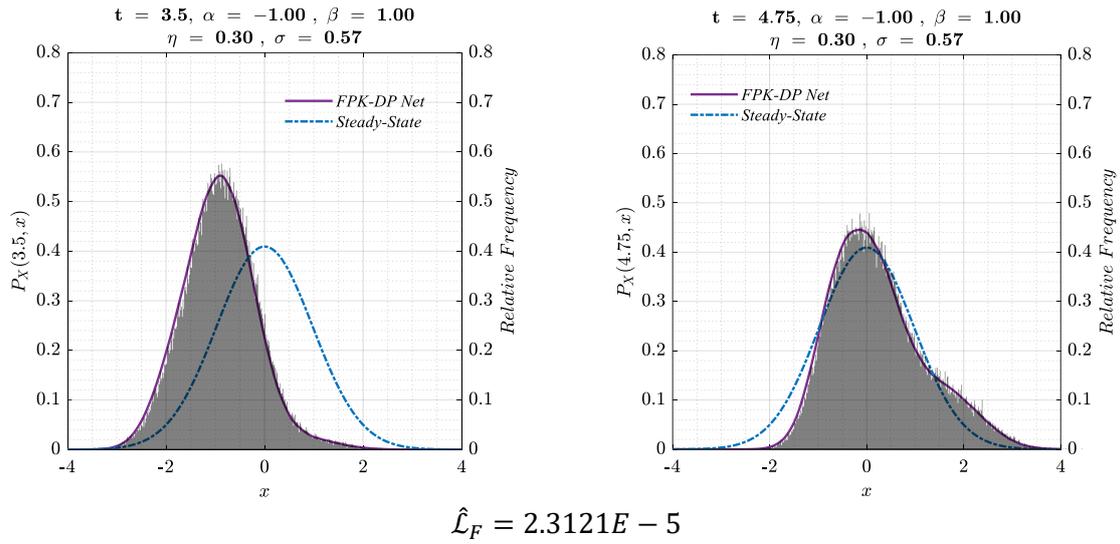

$$\hat{\mathcal{L}}_F = 2.3121E-5$$

Figure 15 – Comparison of the FPK-DP Net predicted PDF with Monte-Carlo simulation results for autonomous Duffing oscillator, where $\alpha = -1$, $\beta = 1$, $\eta = 0.3$, $\gamma = 0$, and $\sigma = 0.57$

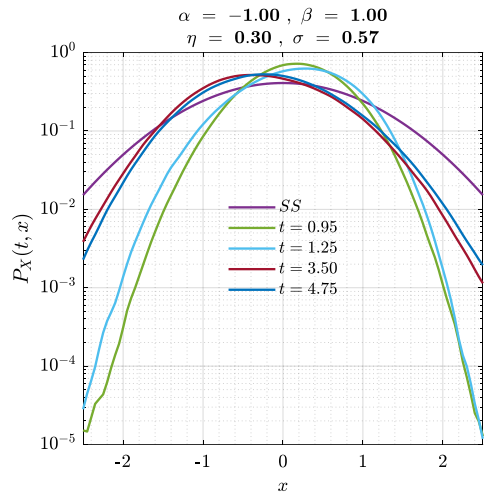

Figure 16 – Evolution of the FPK-DP network output for the autonomous Duffing oscillator in the logarithmic scale

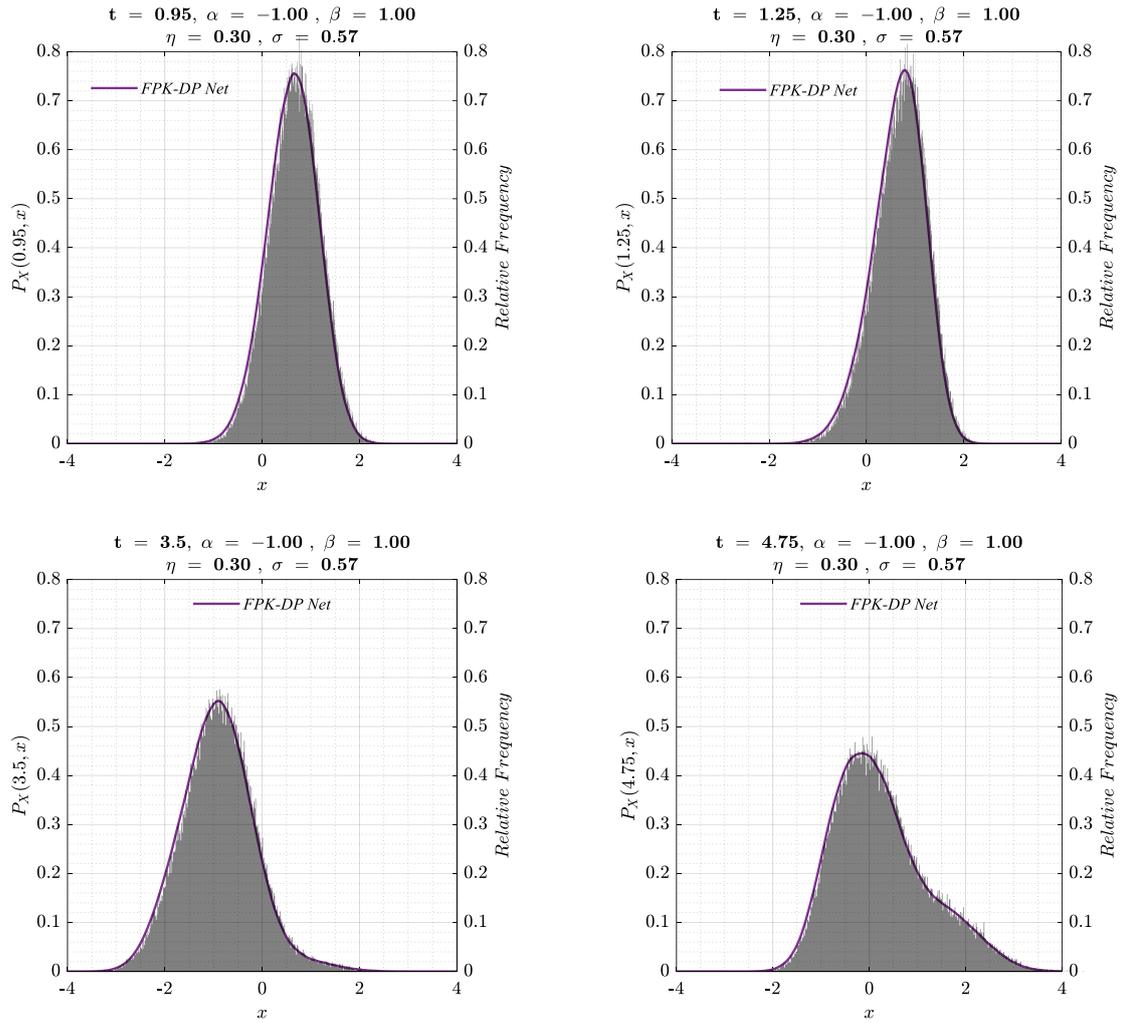

$$\hat{\mathcal{L}}_F = 8.3332E-5$$

Figure 17 – Comparison of the FPK-DP Net predicted PDF with Monte-Carlo simulation results for non-autonomous Duffing oscillator, where $\alpha = -1, \beta = 1, \eta = 0.3, \gamma = 0.65, \omega = 1.2$, and $\sigma = 0.57$

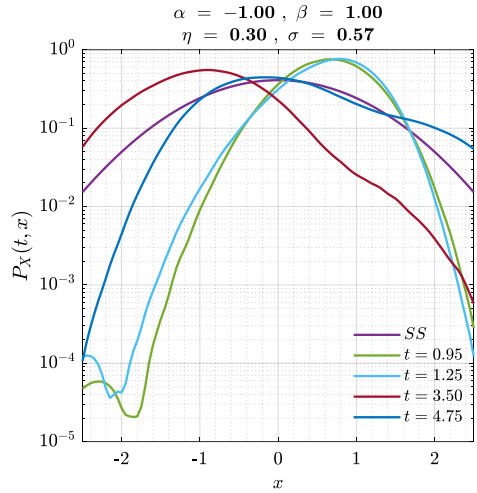

Figure 18 – Evolution of the FPK-DP network output for the non-autonomous Duffing oscillator in the logarithmic scale

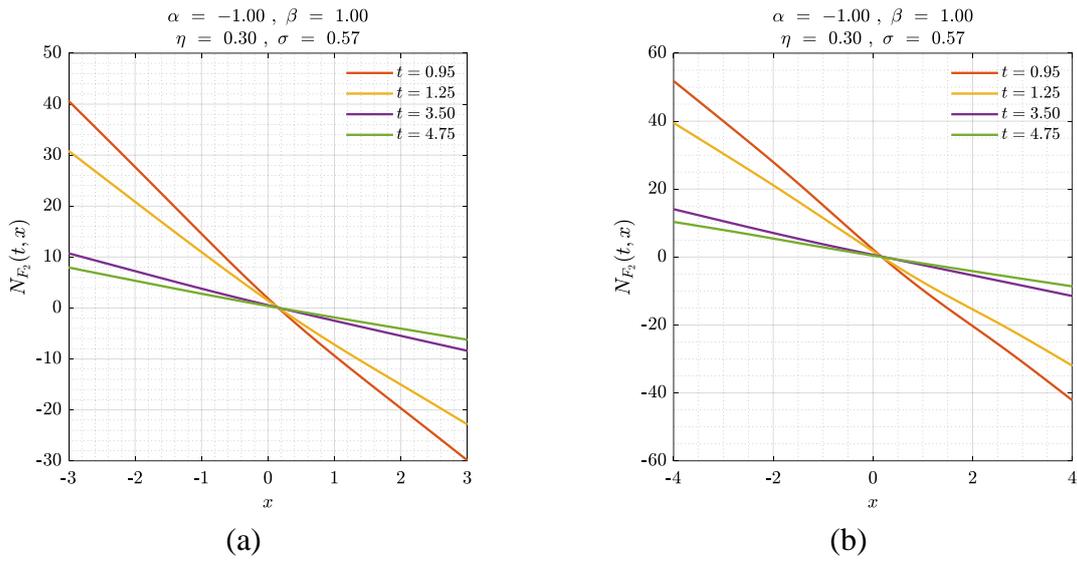

(a)        (b)

Figure 19 – Comparison of the Evolution of the output of the $N_{F_2}(t,x)$ network for (a) the autonomous and (b) the non-autonomous Duffing oscillator

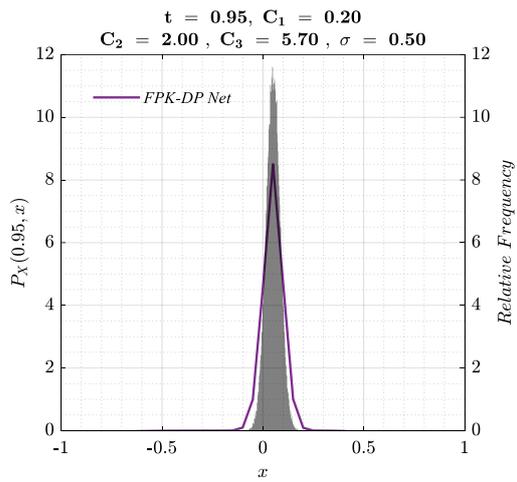
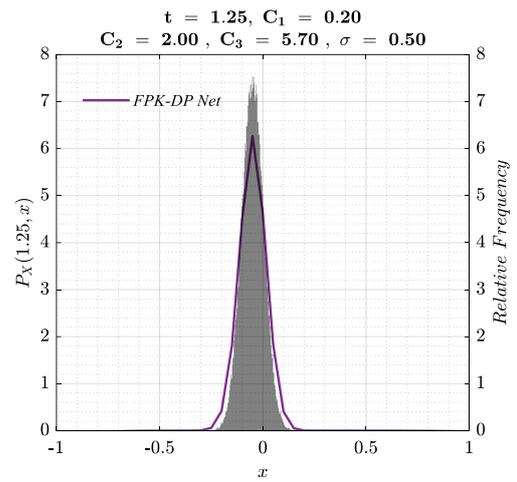
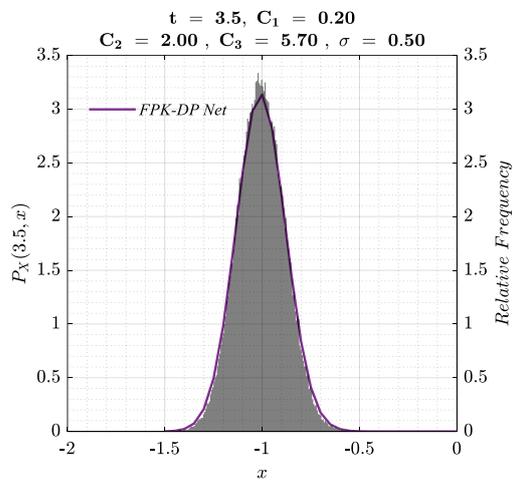
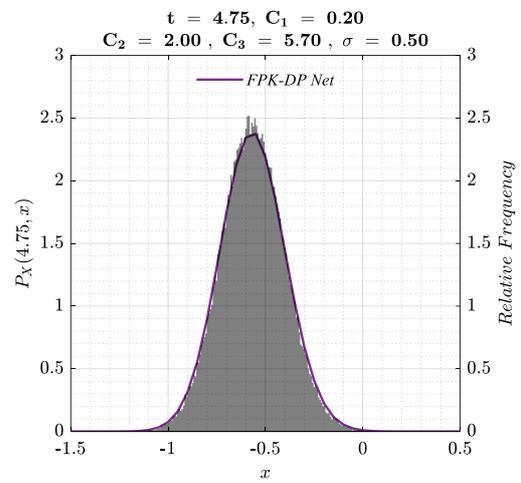

$$\hat{\mathcal{L}}_F = 3.99456E - 4$$

Figure 20 – Comparison of the FPK-DP Net predicted PDF with Monte-Carlo simulation results for the Roessler attractor, where $C_1 = 0.2$, $C_2 = 2$, $C_3 = 5.7$, and $\sigma = 0.5$

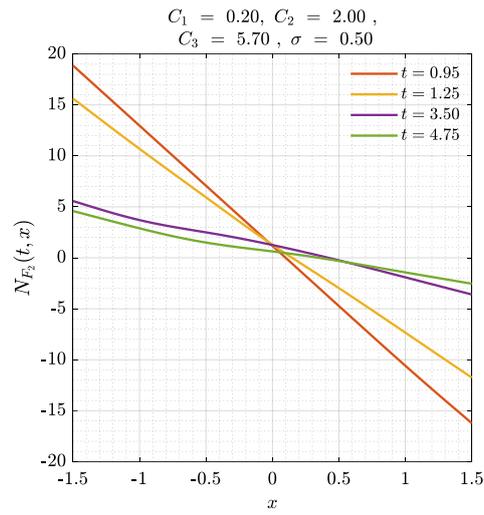

Figure 21 – Evolution of the output of the $N_{F_2}(t,x)$ network for the Roessler attractor